%% file: main.tex
\documentclass[10pt,twocolumn,letterpaper]{article}

\usepackage{cvpr}      %
\usepackage[accsupp]{axessibility}  %
\input{macros}
\usepackage{amsmath}
\usepackage{amssymb}
\usepackage[utf8]{inputenc} %
\usepackage[T1]{fontenc}    %
\usepackage{url}            %
\usepackage{booktabs}       %
\usepackage{amsfonts}       %
\usepackage{nicefrac}       %
\usepackage{microtype}      %
\usepackage{xcolor}         %
\usepackage{graphicx}
\usepackage{enumitem}
\usepackage{array}
\usepackage{booktabs}
\usepackage{multirow}
\usepackage{arydshln}
\usepackage{amsfonts,bm,relsize,dsfont}
\usepackage{booktabs}
\usepackage{caption}

\usepackage[pagebackref=true,breaklinks=true,letterpaper=true,colorlinks,citecolor=citecolor,bookmarks=false]{hyperref}
\definecolor{citecolor}{RGB}{30,102,235}

\DeclareMathOperator*{\argmin}{arg\,min}
\newcommand\blfootnote[1]{%
  \begingroup
  \renewcommand\thefootnote{}\footnote{#1}%
  \addtocounter{footnote}{-1}%
  \endgroup
}

\usepackage[capitalize]{cleveref}
\crefname{section}{Sec.}{Secs.}
\Crefname{section}{Section}{Sections}
\Crefname{table}{Table}{Tables}
\crefname{table}{Tab.}{Tabs.}

\def\cvprPaperID{***}
\def\confName{CVPR}
\def\confYear{2022}

\begin{document}

\title{GAN-Supervised Dense Visual Alignment}

\author{
 \hspace{-6mm}William Peebles$^{1}$ \hspace{2.5mm} Jun-Yan Zhu$^{2}$ \hspace{2.5mm} Richard Zhang$^3$ \hspace{2.5mm}
 Antonio Torralba$^4$ \hspace{2.5mm} Alexei A. Efros$^1$ \hspace{2.5mm} Eli Shechtman$^3$ \\[2mm]
 $^{1}$UC Berkeley \qquad $^2$Carnegie Mellon University \qquad $^3$Adobe Research \qquad $^4$MIT CSAIL \\[2mm]
 Facebook AI Research (FAIR)\vspace{-4mm}
}
\twocolumn[{%
\renewcommand\twocolumn[1][]{#1}%
\maketitle
\vspace{-14mm}
\begin{center}
    \centering
    \captionsetup{type=figure}
    \includegraphics[width=\linewidth]{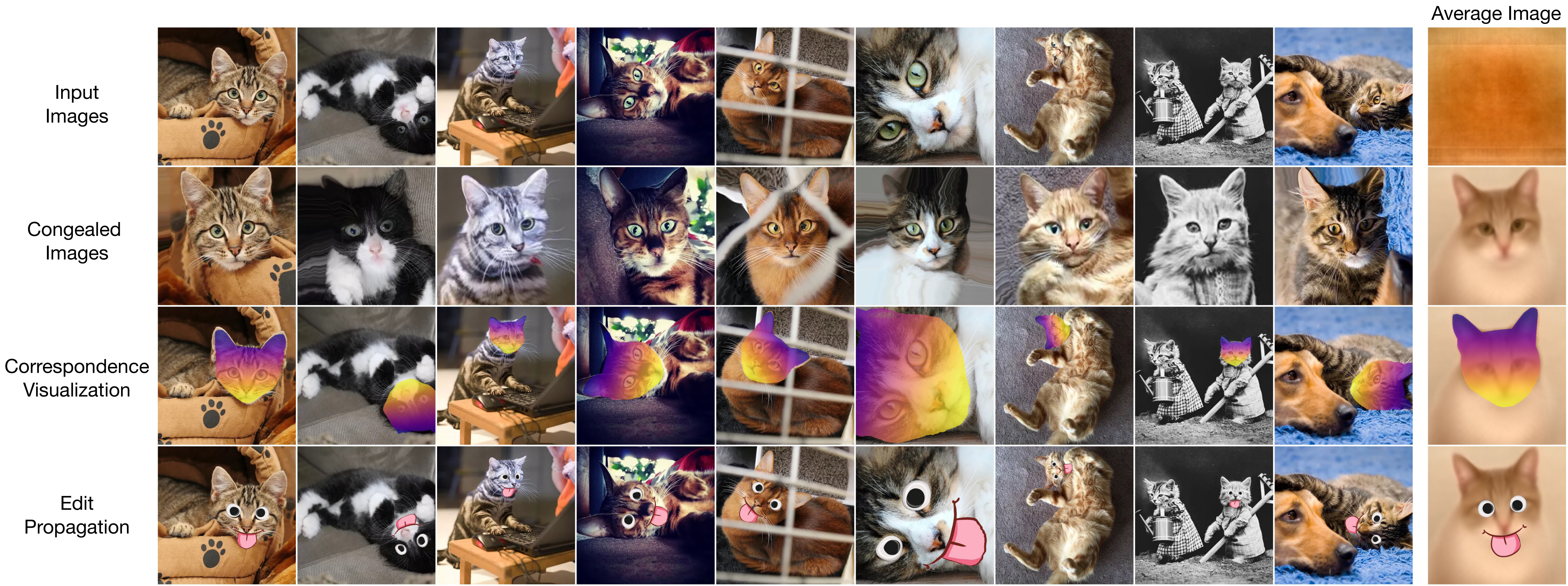}
    \captionof{figure}{Given an input dataset of unaligned images, our GANgealing algorithm discovers dense correspondences between all images. \textbf{Top row}: Images from LSUN Cats and the dataset's average image. \textbf{Second row}: Our learned transformations of the input images. \textbf{Third row}: Dense correspondences learned by GANgealing. \textbf{Bottom row}: By annotating the average transformed image, we can propagate user edits to images and videos. \textbf{Please see our project page for detailed video results:} \url{www.wpeebles.com/gangealing}.}\label{fig:teaser}
\end{center}%
}]

\begin{abstract}\blfootnote{\vspace{-4mm}Code and models: \url{www.github.com/wpeebles/gangealing}}\vspace{-6mm}

We propose GAN-Supervised Learning, a framework for learning discriminative models and their GAN-generated training data jointly end-to-end. We apply our framework to the dense visual alignment problem. Inspired by the classic Congealing method, our GANgealing algorithm trains a Spatial Transformer to map random samples from a GAN trained on unaligned data to a common, jointly-learned target mode. We show results on eight datasets, all of which demonstrate our method successfully aligns complex data and discovers dense correspondences. GANgealing significantly outperforms past self-supervised correspondence algorithms and performs on-par with (and sometimes exceeds) state-of-the-art supervised correspondence algorithms on several datasets---without making use of any correspondence supervision or data augmentation and despite being trained exclusively on GAN-generated data. For precise correspondence, we improve upon state-of-the-art supervised methods by as much as $3\times$. We show applications of our method for augmented reality, image editing and automated pre-processing of image datasets for downstream GAN training.

\end{abstract}

\section{Introduction}
\input{paper_sections/intro}

\section{Related Work}
\input{paper_sections/related_work}

\section{GAN-Supervised Learning}
\input{paper_sections/methods}
\section{Experiments}

\input{paper_sections/experiments}

\section{\vspace{-2mm}Limitations and Discussion}
\input{paper_sections/conclusion}
{\small
\bibliographystyle{ieee_fullname}
\bibliography{egbib}
}
\clearpage
\input{paper_sections/supplementary}

\end{document}

%% file: macros.tex
\newcommand{\myparagraph}[1]{\vspace{-4pt}\paragraph{#1}}

\newcommand{\w}{\mathbf{w}}
\newcommand{\g}{\mathbf{g}}
\newcommand{\p}{\mathbf{c}}
\newcommand{\R}{\mathbb{R}}
\newcommand{\flow}{T_{\text{flow}}}
\newcommand{\simi}{T_{\text{sim}}}
\def\x{\bm{x}} %
\def\y{\bm{y}} %

%% file: paper_sections/intro.tex
Visual alignment, also known as the correspondence or registration problem, is a critical element in much of computer vision, including optical flow, 3D matching, medical imaging, tracking and augmented reality. While much recent progress has been made on pairwise alignment (aligning image A to image B)~\cite{rocco2017convolutional,rocco2018end,rocco2018neighbourhood,Liu_2020_CVPR,seo2018attentive,min2019hyperpixel,min2020learning,cho2021semantic,aberman2018neural,Min_2021_CVPR,dosovitskiy2015flownet,ilg2017flownet,teed2020raft}, the problem of global joint alignment (aligning \textit{all} images across a dataset) has not received as much attention. Yet, joint alignment is crucial for tasks requiring a common reference frame, such as automatic keypoint annotation, augmented reality or edit propagation (see Figure~\ref{fig:teaser} bottom row).  There is also evidence that training on jointly aligned datasets (such as FFHQ~\cite{karras2019style}, AFHQ~\cite{choi2020starganv2}, CelebA-HQ~\cite{karras2018progressive}) can produce higher quality generative models than training on unaligned data.

In this paper, we take inspiration from a series of classic works on automatic joint image set alignment. In particular, we are motivated by the seminal unsupervised {\em Congealing} method of Learned-Miller~\cite{learned-miller2006congealing} which showed that a set of images could be brought into alignment by continually warping them toward a common, updating mode. While Congealing can work surprisingly well on simple binary images, such as MNIST digits, the direct pixel-level alignment is not powerful enough to handle most datasets with significant appearance and pose variation.

To address these limitations, we propose GANgealing: a \textit{GAN-Supervised} algorithm that learns transformations of input images to bring them into better joint alignment. The key is in employing the latent space of a GAN (trained on the unaligned data) to automatically generate paired training data for a Spatial Transformer~\cite{jaderberg2015spatial}. Crucially, in our proposed GAN-Supervised Learning framework, \textit{both} the Spatial Transformer and the target images are learned jointly. Our Spatial Transformer is trained \textit{exclusively} with GAN images and generalizes to real images at test time.

We show results spanning eight datasets---LSUN Bicycles, Cats, Cars, Dogs, Horses and TVs~\cite{yu2015lsun}, In-The-Wild CelebA~\cite{liu2015faceattributes} and CUB~\cite{WahCUB_200_2011}---that demonstrate our GANgealing algorithm is able to discover accurate, dense correspondences across datasets. We show our Spatial Transformers are useful in image editing and augmented reality tasks. Quantitatively, GANgealing significantly outperforms past self-supervised dense correspondence methods, nearly doubling key point transfer accuracy (PCK~\cite{andriluka20142d}) on many SPair-71K~\cite{min2019spair} categories. Moreover, GANgealing sometimes matches and even exceeds state-of-the-art correspondence-\textit{supervised} methods.

%% file: paper_sections/related_work.tex
\myparagraph{Pre-Trained GANs for Vision.} Prior work has explored the use of GANs~\cite{goodfellow2014GAN,radford2015unsupervised} in vision tasks such as classification~\cite{chai2021ensembling,mao2021generative,besnier2020dataset,tanaka2019data,wu2018conditional}, segmentation~\cite{voynov2020big,melas2021finding,zhang2021datasetgan,tritrong2021repurposing} and representation learning~\cite{donahue2016adversarial,donahue2019large,jahanian2021generative,dumoulin2016adversarially,baradad2021learning}, as well as 3D vision and graphics tasks~\cite{shi2021lifting,pan2020gan2shape,zhang2020image,hao2021GANcraft}. Likewise, we share the goal of leveraging the power of pre-trained deep generative models for vision tasks. However, the relevant past methods follow a common two-stage paradigm of (1) synthesizing a GAN-generated dataset and (2) training a discriminative model on the fixed dataset. In contrast, our GAN-Supervised Learning approach learns \textit{both} the discriminative model as well as the GAN-generated data jointly end-to-end. We do not rely on hand-crafted pixel space augmentations~\cite{chai2021ensembling,jahanian2021generative}, human-labeled data~\cite{zhang2021datasetgan,zhang2020image,tritrong2021repurposing,shi2021lifting,hao2021GANcraft} or post-processing of GAN-generated datasets using domain knowledge~\cite{voynov2020big,melas2021finding,zhang2020image,besnier2020dataset}.

\vspace{-2mm}

\myparagraph{Joint Image Set Alignment.} 
Average images have long been used to visualize joint
alignment of image sets of the same semantic content (e.g., \cite{torralba2001average,zhu2014averageExplorer}), with the seminal work of Congealing~\cite{learned-miller2006congealing, huang2007congealing} establishing
unsupervised joint alignment as a research problem.
Congealing uses sequential optimization to gradually minimize
the entropy of the intensity distribution of a set of images by continuously warping each image via a parametric transformation (e.g., affine). It produces impressive results on well-structured datasets, such as digits, but struggles on more complex data. Subsequent work assumes the data lies on a low-rank subspace~\cite{peng2012rasl,kemelmacher2012collection} or factorizes images as a composition of color, appearance and shape~\cite{mobahi2014compositional} to establish dense correspondences between instances of the
same object category.
FlowWeb~\cite{zhou2015flowweb} %
uses cycle consistency constraints to estimate a fully-connected correspondence flow graph. 
Every method above assumes that it is possible to align all images to a single central mode in the data. Joint visual alignment and clustering was proposed in  AverageExplorer~\cite{zhu2014averageExplorer} but as a user-driven data interaction tool. Bounding box supervision has been used to align and cluster multiple modes within object categories~\cite{divvala2012object}.
Automated transformation-invariant clustering methods~\cite{frey1999estimating,frey2003clustering,mattar2012congealcluster} can align images in a collection before comparing them but work only in limited domains. 
Recently, Monnier et al.~\cite{monnier2020deepcluster} showed that warps could be predicted with a network instead, removing the need for per-image optimization; this opened the door for simultaneous alignment and clustering of large-scale collections. Unlike our approach, these methods assume images can be aligned with simple (e.g., affine) color transformations; this assumption breaks down for complex datasets like LSUN.

\vspace{-2mm}

\myparagraph{Spatial Transformer Networks (STNs).}
A Spatial Transformer module~\cite{jaderberg2015spatial} is one way to incorporate learnable geometric transformations in a deep learning framework. It regresses a set of warp parameters, where the warp and grid sampling functions are differentiable to enable backpropagation. STNs have seen success in discriminative tasks (e.g., classification) and applications such as robust filter learning~\cite{dai2017deformable,jia2016dynamic}, view synthesis~\cite{zhou2016view,ganin2016deepwarp,park2017transformation} and 3D representation learning~\cite{kanazawa2016warpnet,yan2016perspective,zhou2017unsupervised}.
Inverse Compositional STNs (IC-STNs)~\cite{lin2017inverse} advocate an iterative image alignment framework in the spirit of the classical Lukas-Kanade algorithm~\cite{lucas1981iterative,baker2004lucas}. Prior work has incorporated STNs in generative models for geometry-texture disentanglement~\cite{xing2018deformable} and image compositing~\cite{lin2018st}. In contrast, we use a generative model to directly produce training data \textit{for} STNs.

%% file: paper_sections/methods.tex
\begin{figure*}[t]
    \centering
    \includegraphics[width=\linewidth]{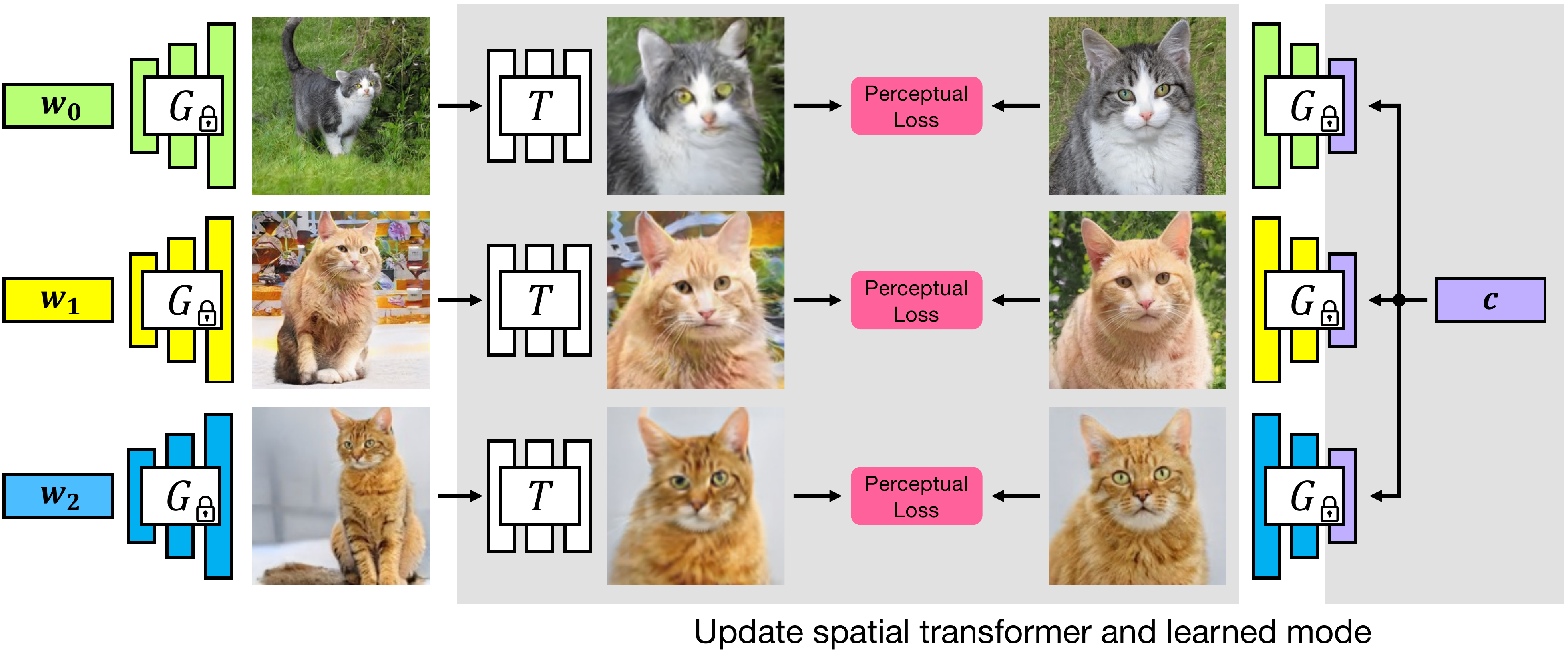}
    \caption{\textbf{GANgealing Overview}. We first train a generator $G$ on unaligned data. We create a \textit{synthetically-generated} dataset for alignment by learning a mode $\p$ in the generator's latent space. We use this dataset to train a Spatial Transformer Network $T$ to map from unaligned to corresponding aligned images using a perceptual loss~\cite{johnson2016perceptual}. The Spatial Transformer generalizes to align \textit{real} images automatically.}
    \label{fig:gangealing}
\end{figure*}
In this section, we present GAN-Supervised Learning. Under this framework, $(\x$, $\y)$ pairs are sampled from a pre-trained GAN generator, where $\x$ is a random sample from the GAN and $\y$ is the sample obtained by applying a \textit{learned} latent manipulation to $\x$'s latent code. These pairs are used to train a network $f_{\theta}: \x \rightarrow \y$. This framework minimizes the following loss:
\begin{align}\label{gan_supervised}
    \mathcal{L}(f_{\theta}, \y) = \ell(f_{\theta}(\x), \y),
\end{align}
where $\ell$ is a reconstruction loss. In vanilla supervised learning, $f_{\theta}$ is learned on \textit{fixed} $(\x, \y)$ pairs. In contrast, in GAN-Supervised Learning, \textit{both} $f_{\theta}$ and the targets $\y$ are learned jointly end-to-end. At test time, we are free to evaluate $f_{\theta}$ on \textit{real} inputs.

\subsection{Dense Visual Alignment}

Here, we show how GAN-Supervised Learning can be applied to Congealing~\cite{learned-miller2006congealing}---a classic unsupervised alignment algorithm. In this instantiation, $f_\theta$ is a Spatial Transformer Network~\cite{jaderberg2015spatial} $T$, and we describe our parameterization of inputs $\x$ and learned targets $\y$ below. We call our algorithm \textit{GANgealing}. We present an overview in Figure~\ref{fig:gangealing}.

GANgealing begins by training a latent variable generative model $G$ on an unaligned input dataset. We refer to the input latent vector to $G$ as $\w \in \R^{512}$. With $G$ trained, we are free to draw samples from the unaligned distribution by computing $\x = G(\w)$ for randomly sampled $\w \sim \mathcal{W}$, where $\mathcal{W}$ denotes the distribution over latents. Now, consider a fixed latent vector $\p \in \R^{512}$. This vector corresponds to a fixed synthetic image $G(\p)$ from the original unaligned distribution. A simple idea in the vein of traditional Congealing is to use $G(\p)$ as the target mode $\y$---i.e., we learn a Spatial Transformer $T$ that is trained to warp every random unaligned image $\x=G(\w)$ to the same target image $\y=G(\p)$. Since $G$ is differentiable in its input, we can optimize $\p$ and hence learn the target we wish to congeal towards. Specifically, we can optimize the following loss with respect to both $T$'s parameters and the target image's latent vector $\p$ jointly:

\begin{align}\label{gangeal_loss}
    \mathcal{L}_{\textrm{align}}(T, \p) = \ell(T(G(\w)), G(\p)),
\end{align}

\noindent where $\ell$ is some distance function between two images.  By minimizing $\mathcal{L}$ with respect to the target latent vector $\p$, GANgealing encourages $\p$ to find a pose that makes $T$'s job as easy as possible. If the current value of $\p$ corresponds to a pose that cannot be reached from most images via the transformations predicted by $T$, then it can be adjusted via gradient descent to a different vector that is ``reachable" by more images. %

This simple approach is reasonable for datasets with limited diversity; however, in the presence of significant appearance and pose variation, it is not reasonable to expect that every unaligned sample can be aligned to the exact same target image. Hence, optimizing the above loss does not produce good results in general (see Table~\ref{pck-abl}). Instead of using the same target $G(\p)$ for every randomly sampled image $G(\w)$, it would be ideal if we could construct a \textit{per-sample target} that retains the appearance of $G(\w)$ but where the pose and orientation of the object in the target image is roughly identical across targets. To accomplish this, given $G(\w)$, we produce the corresponding target by setting just a portion of the $\w$ vector equal to the target vector $\p$. Specifically, let mix$(\p, \w) \in \R^{512}$ refer to the latent vector whose first entries are taken from $\p$ and remaining entries are taken from $\w$. By sampling new $\w$ vectors, we can create an infinite pool of paired data where the input is the unaligned image $\x=G(\w)$ and the target $\y=G(\text{mix}(\p, \w))$ shares the appearance of $G(\w)$ but is in a learned, fixed pose. This gives rise to the GANgealing loss function:

\begin{align}\label{gangeal_loss2}
    \mathcal{L}_{\textrm{align}}(T, \p) = \ell(T(\underbrace{G(\w)}_{\x}), \underbrace{G(\text{mix}(\p, \w))}_{\y}),
\end{align}
where $\ell$ is a perceptual loss function~\cite{johnson2016perceptual}. In this paper, we opt to use StyleGAN2~\cite{karras2020analyzing} as our choice of $G$, but in principle other GAN architectures could be used with our method. An advantage of using StyleGAN2 is that it possesses some innate style-pose disentanglement that we can leverage to construct the per-image target described above. Specifically, we can construct the per-sample targets $G(\text{mix}(\p,\w))$ by using style mixing~\cite{karras2019style}---$\p$ is supplied to the first few inputs to the synthesis generator that roughly control pose and $\w$ is fed into the later layers that roughly control texture. See Table~\ref{pck-abl} for a quantitative ablation of the mixing "cutoff point" where we begin to feed in $\w$  (i.e., the cutoff point is chosen as a layer index in $\mathcal{W}^+$ space~\cite{abdal2019image2stylegan}).

\myparagraph{Spatial Transformer Parameterization.} Recall that a Spatial Transformer $T$ takes as input an image and regresses and applies a (reverse) sampling grid $\g \in \R^{H \times W \times 2}$ to the input image. Hence, one must choose how to constrain the $\g$ regressed by $T$. In this paper, we explore a $T$ that performs similarity transformations (rotation, uniform scale, horizontal shift and vertical shift). We also explore an arbitrarily expressive $T$ that directly regresses unconstrained per-pixel flow fields $\g$. Our final $T$ is a composition of the similarity Spatial Transformer into the unconstrained Spatial Transformer, which we found worked best. In contrast to prior work~\cite{lin2018st,monnier2020deepcluster}, we do not find multi-stage training necessary and train our composed $T$ end-to-end. Finally, our Spatial Transformer is also capable of performing horizontal flips at test time---please refer to Supplement~\ref{apdx:flip} for details.

When using the unconstrained $T$, it can be beneficial to add a total variation regularizer that encourages the predicted flow to be smooth to mitigate degenerate solutions: $\mathcal{L}_{\textrm{TV}}(T) = \mathcal{L}_{\textrm{Huber}}(\Delta_x \g) + \mathcal{L}_{\textrm{Huber}}(\Delta_y \g)$, where $\mathcal{L}_{\textrm{Huber}}$ denotes the Huber loss and $\Delta_x$ and $\Delta_y$ denote the partial derivative w.r.t. $x$ and $y$ coordinates under finite differences. We also use a regularizer that encourages the flow to not deviate from the identity transformation: $\mathcal{L}_{\textrm{I}}(T) = ||\g||_2^2$.

\myparagraph{Parameterization of \texorpdfstring{$\p$}{c}.} In practice, we do not backpropagate gradients directly into $\p$. Instead, we parameterize $\p$ as a linear combination of the top-$N$ principal directions of $\mathcal{W}$ space~\cite{harkonen2020ganspace,tian2021a}:
\begin{align}
    \p = \bar{\w} + \sum_{i=1}^N \alpha_i \mathbf{d}_i,
\end{align}
where $\bar{\w}$ is the empirical mean $\w$ vector, $\mathbf{d}_i$ is the $i$-th principal direction and $\alpha_i$ is the learned scalar coefficient of the direction. Instead of optimizing $\mathcal{L}$ w.r.t. $\p$ directly, we optimize it w.r.t. the coefficients $\{\alpha_i\}_{i=1}^N$. The motivation for this reparameterization is that StyleGAN's $\mathcal{W}$ space is highly expressive. Hence, in the absence of additional constraints, naive optimization of $\p$ can yield poor target images off the manifold of natural images. Decreasing $N$ keeps $\p$ on the manifold and prevents degenerate solutions. See Table~\ref{pck-abl} for an ablation of $N$.

Our final GANgealing objective is given by:

\vspace{-4mm}
\begin{multline}
    \mathcal{L}(T,\p) = \mathbb{E}_{\w \sim \mathcal{W}}[\mathcal{L}_{\textrm{align}}(T,\p) \\+ \lambda_{\textrm{TV}} \mathcal{L}_{\textrm{TV}}(T) + \lambda_{I} \mathcal{L}_\textrm{I}(T)].
\end{multline}

\noindent We set the loss weighting $\lambda_{\textrm{TV}}$ at either $1000$ or $2500$ (depending on choice of $\ell$) and the loss weighting $\lambda_I$ at 1. See Supplement~\ref{apdx:implementation} for additional details and hyperparameters.

\subsection{Joint Alignment and Clustering}\label{sec:clustering}

GANgealing as described so far can handle highly-multimodal data (e.g., LSUN Bicycles, Cats, etc.). Some datasets, such as LSUN Horses, feature extremely diverse poses that cannot be represented well by a single mode in the data. To handle this situation, GANgealing can be adapted into a clustering algorithm by simply learning more than one target latent $\p$. Let $K$ refer to the number of $\p$ vectors (clusters) we wish to learn. Since each $\p$ captures a specific mode in the data, learning multiple $\{\p_k\}_{k=1}^K$ would enable us to learn multiple modes. Now, each $\p_k$ will learn its own set of $\mathbf{\alpha}$ coefficients. Similarly, we will now have $K$ Spatial Transformers, one for each mode being learned. This variant of GANgealing amounts to simultaneously clustering the data and learning dense correspondence between all images within each cluster. To encourage each $\p_k$ and $T_k$ pair to specialize in a particular mode, we include a hard-assignment step to assign unaligned synthetic images to modes:
\begin{align}
    \mathcal{L}_{\textrm{align}}^K(T, \p) = \min_k \mathcal{L}_{\textrm{align}}(T_k, \p_k)
\label{eq:assignment}
\end{align}
Note that the $K=1$ case is equivalent to the previously described unimodal case. At test time, we can assign an input fake image $G(\w)$ to its corresponding cluster index $k^* = \argmin_k \mathcal{L}_{\textrm{align}}(T_k, \p_k)$. Then, we can warp it with the Spatial Transformer $T_{k^*}$. However, a problem arises in that we cannot compute this cluster assignment for input \textit{real} images---the assignment step requires computing $\mathcal{L}_{\textrm{align}}$, which itself requires knowledge of the input image's corresponding $\w$ vector. The most obvious solution to this problem is to perform GAN inversion~\cite{zhu2016generative,brock2017neural,bau2019ganpaint} on input real images $\x$ to obtain a latent vector $\w$ such that $G(\w) \approx \x$. However, accurate GAN inversion for non-face datasets remains somewhat challenging and slow, despite recent progress~\cite{alaluf2021restyle,huh2020ganprojection}. Instead, we opt to train a classifier that directly predicts the cluster assignment of an input image. We train the classifier using a standard cross-entropy loss on (input fake image, target cluster) pairs $(G(\w), k^*)$, where $k^*$ is obtained using the above assignment step. We initialize the classifier with the weights of $T$ (replacing the warp head with a randomly-initialized classification head). As with the Spatial Transformer, the classifier generalizes well to real images despite being trained exclusively on fake samples. 

\begin{figure*}[t!]
    \centering
    \includegraphics[width=\linewidth]{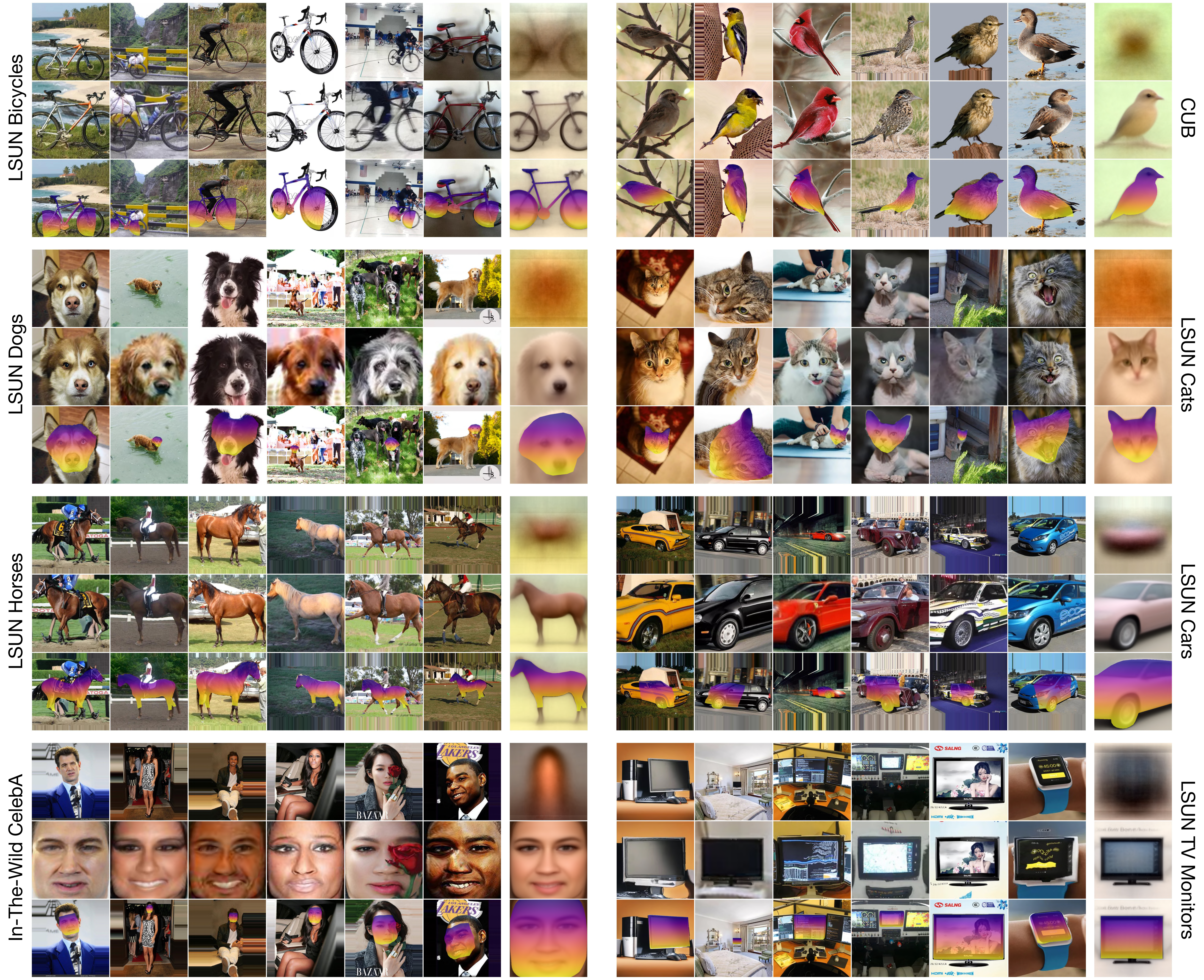}
    \caption{\textbf{Dense correspondence results on eight datasets}. For each dataset, the top row shows unaligned images and the dataset average image. The middle row shows our learned alignment of the input images. The bottom row shows dense correspondences between the images. For our clustering models (LSUN Horses and Cars), we show results for one selected cluster. See Supplement~\ref{apdx:uncurated} for uncurated results.}
    \label{fig:variety}
\end{figure*}

%% file: paper_sections/experiments.tex
In this section, we present quantitative and qualitative results of GANgealing on eight datasets: LSUN Bicycles, Cats, Cars, Dogs, Horses and TVs~\cite{yu2015lsun}, In-The-Wild CelebA~\cite{liu2015faceattributes} and CUB-200-2011~\cite{WahCUB_200_2011}. These datasets feature significant diversity in appearance, pose and occlusion of objects. Only LSUN Cars and Horses use clustering ($K=4$)\footnote{$K$ is a hyperparameter that can be set by the user. We found $K=4$ to be a good default choice for our clustering models.}; for all other datasets we use unimodal GANgealing ($K=1$). Note that all figures except Figure~\ref{fig:gangealing} show our method applied to real images---not GAN samples. Please see \url{www.wpeebles.com/gangealing} for full results.

\begin{figure*}
    \centering
    \vspace{-45mm}
    \includegraphics[width=\linewidth]{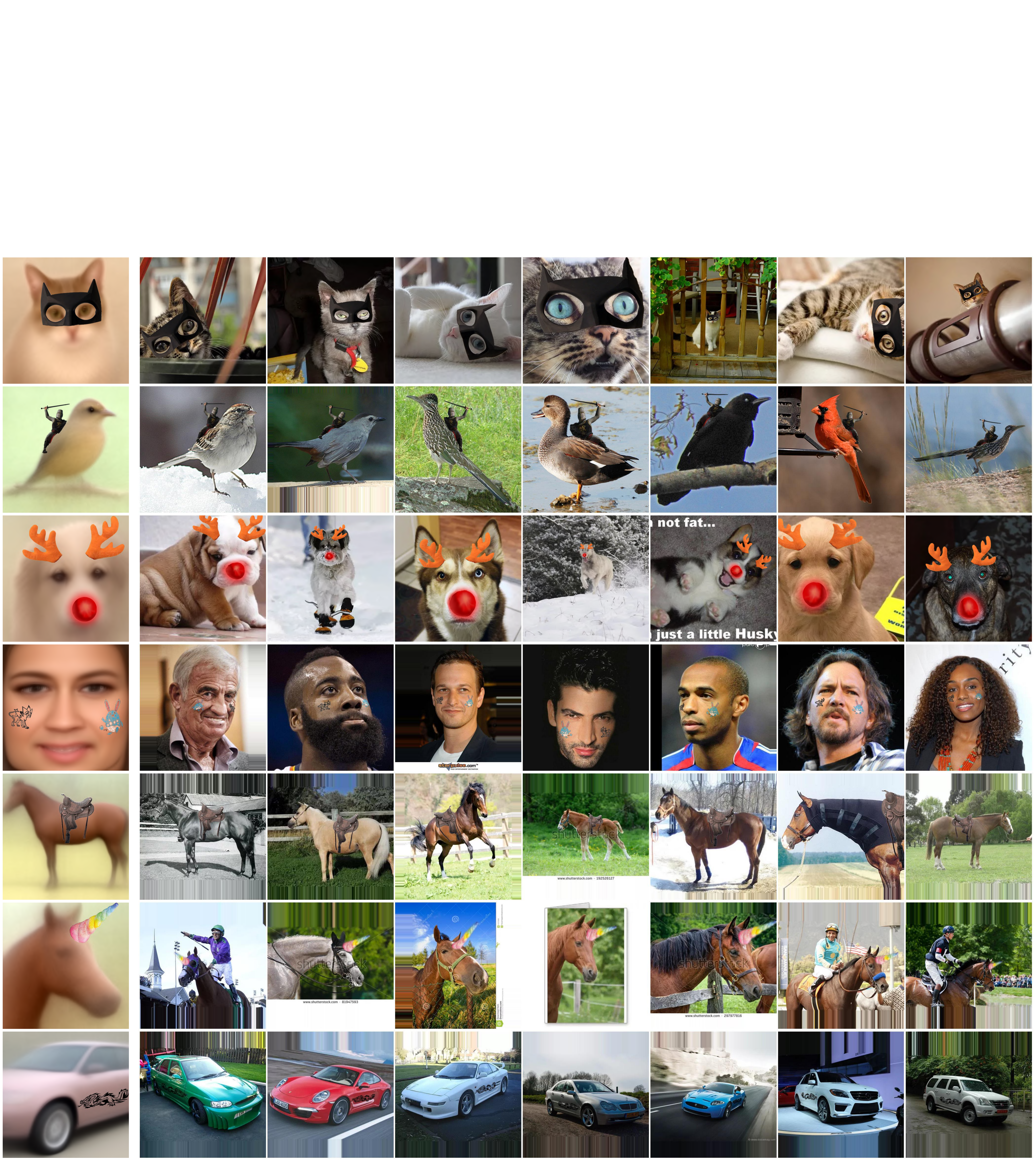}
    \caption{\textbf{Image editing with GANgealing}. By annotating just a single image per-category (our average transformed image), a user can propagate their edits to any image or video in the same category.}
    \label{fig:editing}
    \vspace*{-2mm}
\end{figure*}

\subsection{Propagation from Congealed Space}

With the Spatial Transformer $T$ trained, it is trivial to identify dense correspondences between real input images $\x$. A particularly convenient way to find dense correspondences between a set of images is by propagating from our \textit{congealed coordinate space}. As described earlier, $T$ both regresses and applies a sampling grid $\g$ to an input image. Because we use reverse sampling, this grid tells us where each point in the congealed image $T(\x)$ maps to in the original image $\x$. This enables us to propagate \textit{anything} from the congealed coordinate space---dense labels, sparse keypoints, etc. If a user annotates a \textit{single} congealed image (or the average congealed image) they can then propagate those labels to an entire dataset by simply predicting the grid $\g$ for each image $\x$ in their dataset via a forward pass through $T$. Figures~\ref{fig:teaser} and~\ref{fig:variety} show visual results for all eight datasets---our method can find accurate dense correspondences in the presence of significant appearance and pose diversity. GANgealing accurately handles diverse morphologies of birds, cats with varying facial expressions and bikes in different orientations.

\myparagraph{Image Editing.} Our average congealed image is a template that can propagate any user edit to images of the same category. For example, by drawing cartoon eyes or overlaying a Batman mask on our average congealed cat, a user can effortlessly propagate their edits to massive numbers of cat images with forward passes of $T$. We show editing results on several datasets in Figures~\ref{fig:editing} and ~\ref{fig:teaser}.

\myparagraph{Augmented Reality.} Just as we can propagate dense correspondences to images, we can also propagate to individual video frames. Surprisingly, we find that GANgealing yields remarkably smooth and consistent results when applied \textit{out-of-the-box} to videos per-frame without leveraging any temporal information. This enables mixed reality applications like dense tracking and filters. GANgealing can outperform supervised methods like RAFT~\cite{teed2020raft}---please see \url{www.wpeebles.com/gangealing} for results.

\begin{figure*}[t!]
    \centering
    \includegraphics[width=\linewidth]{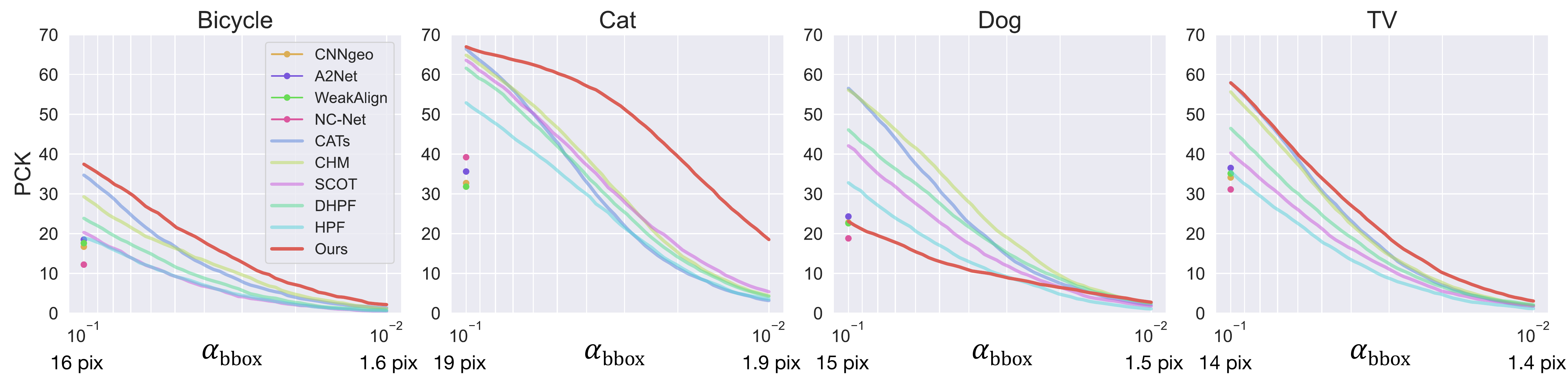}
    \caption{\textbf{PCK@$\alpha_{\text{bbox}}$ on various SPair-71K categories for $\alpha_{\text{bbox}}$ between $10^{-1}$ and $10^{-2}$}. We report the average threshold (maximum distance for a correspondence to be deemed correct) in pixels for 256$\times$256 images beneath each plot. GANgealing outperforms state-of-the-art supervised methods for very precise thresholds ($<2$ pixel error tolerance), sometimes by substantial margins.}
    \label{fig:pck_curves}
\end{figure*}

\subsection{Direct Image-to-Image Correspondence}

In addition to propagating correspondences from congealed space to unaligned images, we can also find dense correspondences directly between any pair of images $\x_A$ and $\x_B$. At a high level, this merely involves applying the forward warp that maps points in $\x_A$ to points in $T(\x_A)$ and composing it with the reverse warp that maps points in the congealed coordinate space back to $\x_B$. Please refer to Supplement~\ref{apdx:transfer} for details.  %

\myparagraph{Quantitative Results.} We evaluate GANgealing with PCK-Transfer. Given a source image $\x_A$, target image $\x_B$ and ground-truth keypoints for both images, PCK-Transfer measures the percentage of keypoints transferred from $\x_A$ to $\x_B$ that lie within a certain radius of the ground-truth keypoints in $\x_B$.

We evaluate PCK on SPair-71K~\cite{min2019spair} and CUB. For SPair, we use the $\alpha_{\text{bbox}}$ threshold in keeping with prior works. Under this threshold, a predicted keypoint is deemed to be correctly transferred if it is within a radius $\alpha_{\text{bbox}}\max(H_{\text{bbox}},W_{\text{bbox}})$ of the ground truth, where $H_{\text{bbox}}$ and $W_{\text{bbox}}$ are the height and width of the object bounding box in the target image. For each SPair category, we train a StyleGAN2 on the corresponding LSUN category\footnote{We use off-the-shelf StyleGAN2 models for LSUN Cats, Dogs and Horses. Note that we do not evaluate PCK on our clustering models (LSUN Cars and Horses) as these models can only transfer points between images in the same cluster.}---the GANs are trained on $256\times256$ center-cropped images. We then train a Spatial Transformer using GANgealing and directly evaluate on SPair. For CUB, we first pre-train a StyleGAN2 with ADA~\cite{karras2020ADA} on the NABirds dataset~\cite{van2015building} and fine-tune it with FreezeD~\cite{mo2020freeze} on the training split of CUB, using the same image pre-processing and dataset splits as ACSM~\cite{kulkarni2020articulation} for a fair comparison. When $T$ performs a horizontal flip for one image in a pair, we permute our model's predictions for keypoints with a left versus right distinction. 

\begin{table}[t]
\centering
    \resizebox{\linewidth}{!}{
    \begin{tabular}{ lccccc } 
    \toprule 
    \multirow{3}{*}{\textbf{Method}} & \multirow{3}{*}{\shortstack[c]{\textbf{Correspondence Supervision}}} & \multicolumn{4}{c}{\textbf{SPair-71K Category}} \\ \cmidrule(lr){3-6}
    & & Bicycle & Cat & Dog & TV \\ \hline
    HPF~\cite{min2019hyperpixel} & matching pairs + keypoints & 18.9 & 52.9 & 32.8 & 35.6 \\ 
    DHPF~\cite{min2020learning} & matching pairs + keypoints & 23.8 & 61.6 & 46.1 & 46.5 \\
    SCOT~\cite{Liu_2020_CVPR} & matching pairs + keypoints* & 20.7 & 63.1 & 42.5 & 40.8 \\
    CHM~\cite{Min_2021_CVPR} & matching pairs + keypoints & 29.3 & 64.9 & 56.1 & 55.6 \\
    CATs~\cite{cho2021semantic} & matching pairs + keypoints & 34.7 & 66.5 & 56.5 & 58.0 \\ \hline
    WeakAlign~\cite{rocco2018end} & matching image pairs & 17.6 & 31.8 & 22.6 & 35.1 \\ 
    NC-Net~\cite{rocco2018neighbourhood} & matching image pairs & 12.2 & 39.2 & 18.8 & 31.1 \\ \hline
    CNNgeo~\cite{rocco2017convolutional} & self-supervised & 16.7 & 32.7 & 22.8 & 34.1 \\
    A2Net~\cite{seo2018attentive} & self-supervised & 18.5 & 35.6 & \textbf{24.3} & 36.5 \\ 
    \textbf{GANgealing} & \textbf{GAN-supervised} & \textbf{37.5} & \textbf{67.0} & 23.1 & \textbf{57.9} \\
    \bottomrule
    \end{tabular}
    }
    \vspace{-2mm}
    \caption{\textbf{PCK-Transfer@$\alpha_{\text{bbox}}=0.1$ results on SPair-71K categories} (test split).\vspace{-2mm}}\label{spair_pck}
\end{table}

\myparagraph{SPair-71K Results.} We compare against several self-supervised and state-of-the-art supervised methods on the challenging SPair-71K dataset in Table~\ref{spair_pck}, using the standard $\alpha_{\text{bbox}}=0.1$ threshold. Our method significantly outperforms prior self-supervised methods on several categories, nearly doubling the best prior self-supervised method's PCK on SPair Bicycles and Cats. \emph{GANgealing performs on par with and even outperforms state-of-the-art correspondence-supervised methods on several categories.} We increase the previous best PCK on Bicycles achieved by Cost Aggregation Transformers~\cite{cho2021semantic} from 34.7\% to 37.5\% and perform comparably on Cats and TVs.

\myparagraph{High-Precision SPair-71K Results.} The usual $\alpha_{\text{bbox}} = 0.1$ threshold reported by most papers using SPair deems a correspondence correct if it is localized within roughly 10 to 20 pixels of the ground truth for $256\times256$ images (depending on the SPair category). In Figure~\ref{fig:pck_curves}, we evaluate performance over a range of thresholds between $0.1$ and $0.01$ (the latter of which affords a roughly 1 to 2 pixel error tolerance, again depending on category). GANgealing outperforms all supervised methods at these high-precision thresholds across all four categories tested. Notably, our LSUN Cats model improves the previous best SPair Cats PCK@$\alpha_{\text{bbox}}=0.01$ achieved by SCOT~\cite{Liu_2020_CVPR} from 5.4\% to 18.5\%. On SPair TVs, we improve the best supervised PCK achieved by Dynamic Hyperpixel Flow~\cite{min2020learning} from 2.1\% to 3.0\%. Even on SPair Dogs, where GANgealing is outperformed by every supervised method at low-precision thresholds, we marginally outperform all baselines at the $0.01$ threshold.

\myparagraph{CUB Results.} Table~\ref{cub_pck} shows PCK results on CUB, comparing against several 2D and 3D correspondence methods that use varying amounts of supervision. GANgealing achieves 57.5\% PCK, outperforming all past methods that require instance mask supervision and performing comparably with the best correspondence-supervised baseline (58.5\%).

\begin{figure}[t]
    \centering
    \includegraphics[width=\linewidth]{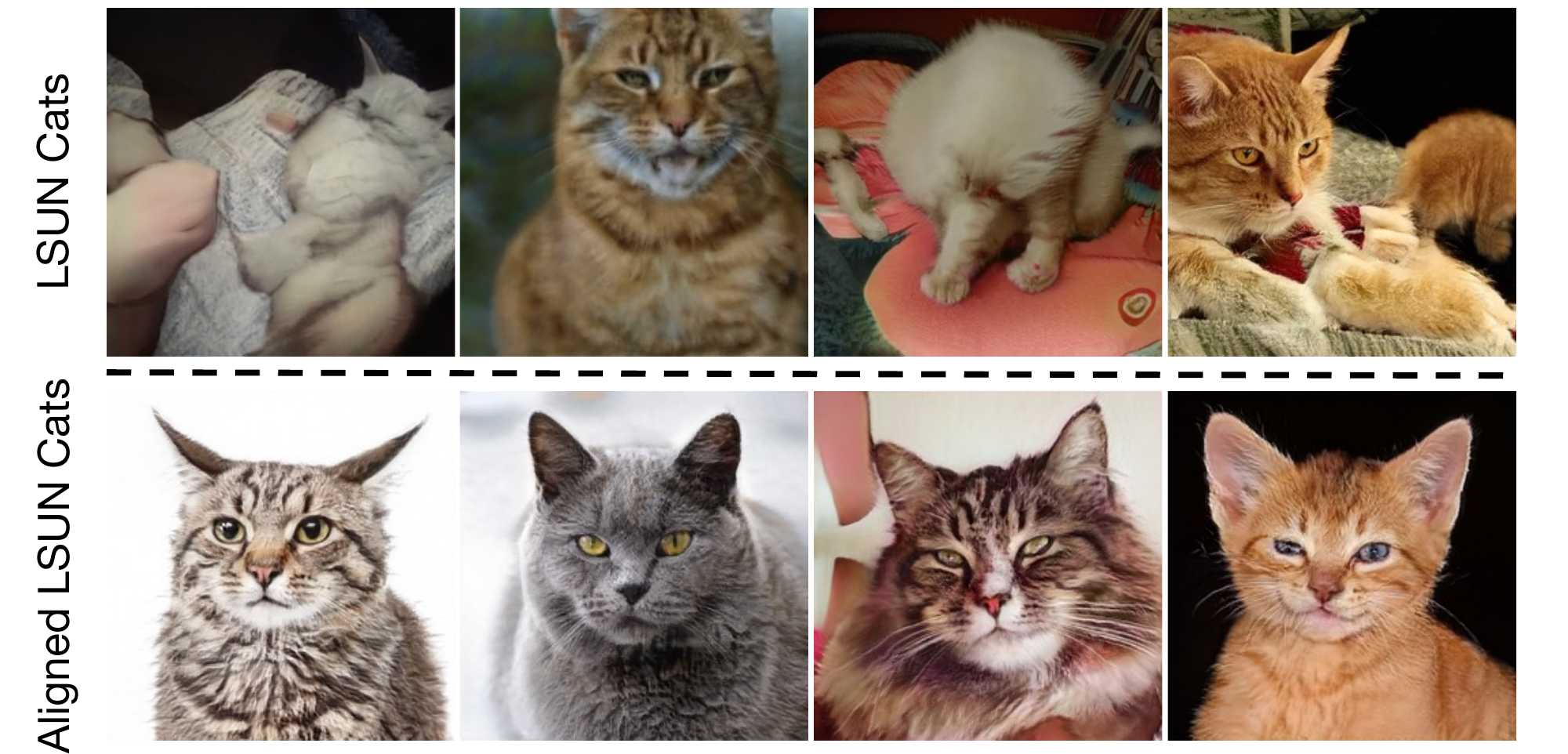}
    \caption{\textbf{GANgealing alignment improves downstream GAN training}. We show random, untruncated samples from StyleGAN2 trained on LSUN Cats versus our aligned LSUN Cats (both models trained from scratch). Our method improves visual fidelity.}\vspace{-2mm}
    \label{fig:gan_samples}
\end{figure}

\begin{figure}
    \centering
    \includegraphics[width=\linewidth]{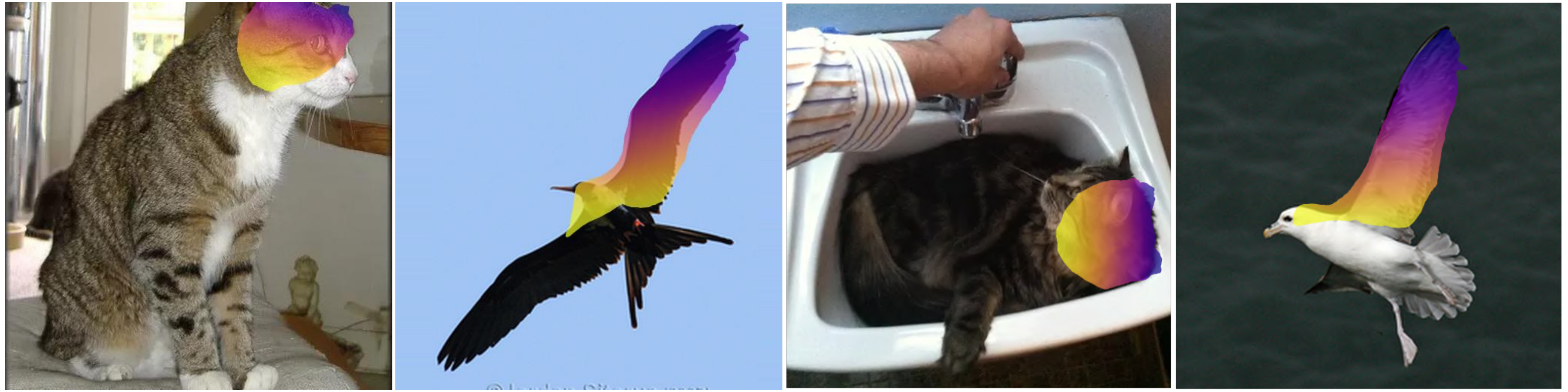}
    \caption{\textbf{Various failure modes}: significant out-of-plane rotation and complex poses poorly modeled by GANs.\vspace{-5mm}}
    \label{fig:failure_cases}
\end{figure}

\myparagraph{Ablations.}\vspace{-2mm} We ablate several components of GANgealing in Table~\ref{pck-abl}. We find that learning the target mode $\p$ is critical for complex datasets; fixing $\p = \bar{\w}$ dramatically degrades PCK from 67\% to 10.6\% for our LSUN Cats model. This highlights the value of our GAN-Supervised Learning framework where \textit{both the discriminative model and targets are learned jointly}. We additionally find that our baseline inspired by traditional Congealing (using a single learned target $G(\p)$ for all inputs) is highly unstable and degrades PCK to as little as 7.7\%. This result demonstrates the importance of our \textit{per-input} alignment targets. We also ablate two choices of the perceptual loss $\ell$: an off-the-shelf supervised option (LPIPS~\cite{zhang2018LPIPS}) and a fully-unsupervised VGG-16~\cite{simonyan2014deep} pre-trained with SimCLR~\cite{chen2020simple} on ImageNet-1K~\cite{deng2009imagenet} (SSL)---there is no significant difference in performance between the two ($\pm 0.2\%$). Please see Table~\ref{pck-abl} for more ablations.

\subsection{Automated GAN Dataset Pre-Processing}
An exciting application of GANgealing is automated dataset pre-processing. Dataset alignment is an important yet costly step for many machine learning methods. GAN training in particular benefits from carefully-aligned and filtered datasets, such as FFHQ~\cite{karras2019style}, AFHQ~\cite{choi2020starganv2} and CelebA-HQ~\cite{karras2018progressive}. We can align input datasets using our similarity Spatial Transformer $T$ to train generators with higher visual fidelity. We show results in Figure~\ref{fig:gan_samples}: training StyleGAN2 from scratch with our learned pre-processing of LSUN Cats yields high-quality samples reminiscent of AFHQ. As we show in Supplement~\ref{accel}, our pre-processing accelerates GAN training significantly.

\begin{table}
    \vspace{0mm}
    \resizebox{\linewidth}{!}{
    \begin{tabular}{ lccc } 
    \toprule 
    \multirow{3}{*}{\textbf{Method}} & \multicolumn{2}{c}{\textbf{Supervision Required}} & \multirow{3}{*}{\shortstack[c]{\textbf{PCK@0.1}}} \\ \cmidrule(lr){2-3}
    & Inst. Mask & Keypoints & \\ \hline
    
    Rigid-CSM (with keypoints)~\cite{kulkarni2019csm} & $\checkmark$ & $\checkmark$ & 45.8 \\
    ACSM (with keypoints)~\cite{kulkarni2020articulation} & $\checkmark$ & $\checkmark$ & 51.0 \\
    IMR (with keypoints)~\cite{tulsiani2020imr} & $\checkmark$ & $\checkmark$ & 58.5 \\\hline

    Dense Equivariance~\cite{thewlis2017unsupervised} & $\checkmark$ & & 33.5 \\ 
    Rigid-CSM~\cite{kulkarni2019csm} & $\checkmark$ & & 36.4 \\ 
    ACSM~\cite{kulkarni2020articulation} & $\checkmark$ & & 42.6 \\ 
    IMR~\cite{tulsiani2020imr} & $\checkmark$ & & 53.4 \\ \hline
    Neural Best Buddies~\cite{aberman2018neural} & & & 35.1 \\ 
    Neural Best Buddies (with flip heuristic) & & & 37.8 \\
    \textbf{GANgealing} & & & \textbf{57.5} \\
    \bottomrule
    \end{tabular}
    }
    \vspace{-2mm}
    \caption{\textbf{PCK-Transfer@$0.1$ on CUB}. Numbers for the 3D methods are reported from~\cite{kulkarni2020articulation}. We sample 10,000 random pairs from the CUB validation split as in~\cite{kulkarni2020articulation}.}\label{cub_pck}
    \vspace{3mm}
    \hfill
    \resizebox{\linewidth}{!}{
    \begin{tabular}{ lcccccc }
    \toprule 
    \textbf{Ablation Description} & \textbf{Loss} ($\ell$) & \textbf{$\mathcal{W}^+$ cutoff} & \textbf{$\lambda_{TV}$} & \textbf{$N$} & \textbf{PCK} \\ \hline
    Don't learn $\p$ (fix $\p = \bar{\w}$) & SSL & 5 & 1000 & 0 & 10.6 \\
    Unconstrained $\p$ optimization & SSL & 5 & 1000 & 512 & 0.34 \\
    Early style mixing cutoff & SSL & 4 & 1000 & 1 & 60.5 \\ 
    Late style mixing cutoff & SSL & 6 & 1000 & 1 & 65.0 \\
    No style mixing & SSL & 14 & 1000 & 1 & 25.9 \\
    No style mixing (LPIPS) & LPIPS & 14 & 1000 & 1 & 7.74 \\
    No $\mathcal{L}_{TV}$ regularizer & SSL & 5 & 0 & 1 & 59.0 \\
    Lower $\lambda_{TV}$ (LPIPS) & LPIPS & 5 & 1000 & 1 & 66.7 \\ \hline
    Complete model (SSL) & SSL & 5 & 1000 & 1 & \textbf{67.2} \\
    Complete model (LPIPS) & LPIPS & 5 & 2500 & 1 & \textbf{67.0} \\
    \bottomrule
    \end{tabular}
    }
    \vspace{-2mm}
    \caption{\textbf{GANgealing ablations for LSUN Cats}. We evaluate on SPair-71K Cats using $\alpha_{\text{bbox}} = 0.1$. SSL refers to using a self-supervised VGG-16 as the perceptual loss $\ell$. $N$ refers to the number of $\mathcal{W}$ space PCA coefficients learned when optimizing $\p$. Note that the LSUN Cats StyleGAN2 generator has 14 layers.\vspace{-3mm}}\label{pck-abl}

\end{table}

%% file: paper_sections/conclusion.tex
Our Spatial Transformer has a few notable failure modes as demonstrated in Figure~\ref{fig:failure_cases}. One limitation with GANgealing is that we can only reliably propagate correspondences that are visible in our learned target mode. For example, the learned mode of our LSUN Dogs model is the upper-body of a dog---this particular model is thus incapable of finding correspondences between, e.g., paws. A potential solution to this problem is to initialize the learned mode with a user-chosen image via GAN inversion that covers all points of interest. Despite this limitation, we obtain competitive results on SPair for some categories where many keypoints are not visible in the learned mode (e.g., cats).

In this paper, we showed that GANs can be used to train highly competitive dense correspondence algorithms from scratch with our proposed GAN-Supervised Learning framework. We hope this paper will lead to increased adoption of GAN-Supervision for other challenging tasks.

\myparagraph{Acknowledgements.} We thank Tim Brooks for his antialiased sampling code; Tete Xiao, Ilija Radosavovic, Taesung Park, Assaf Shocher, Phillip Isola, Angjoo Kanazawa, Shubham Goel, Allan Jabri, Shubham Tulsiani and Dave Epstein for helpful discussions; Rockey Hester for his assistance with testing our model on video. This material is based upon work supported by the National Science Foundation Graduate Research Fellowship Program under Grant No. DGE 2146752. Any opinions, findings, and conclusions or recommendations expressed in this material are those of the authors and do not necessarily reflect the views of the National Science Foundation. Additional funding provided by Berkeley DeepDrive, SAP and Adobe.

%% file: paper_sections/supplementary.tex
\appendix
\twocolumn[{%
\renewcommand\twocolumn[1][]{#1}%
\begin{center}
    {\bf \Large GAN-Supervised Dense Visual Alignment\\[4pt]
    Supplementary Materials}
\end{center}
}]

\setcounter{section}{0}
\renewcommand\thesection{\Alph{section}}

\section{Video Results}
\textbf{Please see our project page at} \url{www.wpeebles.com/gangealing} for animated results, augmented reality applications and visual comparisons against other methods.

\section{Implementation Details}\label{apdx:implementation}

Below we provide details regarding the implementation of GANgealing. Our code is also available at \url{www.github.com/wpeebles/gangealing}.

\subsection{Spatial Transformer}

Our full Spatial Transformer $T$ is composed of two individual Spatial Transformers: $\simi$---a network that takes an image as input and regresses and applies a similarity transformation to the input---and $\flow$---a network that takes an image as input and regresses and applies an unconstrained dense flow field to the input. The warped output image produced by $\simi$ is fed directly into $\flow$. Our final composed Spatial Transformer is given by $T = \flow \circ \simi$, where we compose the warps predicted by the two individual networks and apply it to the original input to obtain the final output.

The architectures of $\simi$ and $\flow$ are identical up to their final layers. Specifically, they each use a ResNet backbone~\cite{he2016deep}, closely following the design of the ResNet-based discriminator from StyleGAN2~\cite{karras2020analyzing}; in particular, we do not use normalization layers. We do \textit{not} use weight sharing for any layers of the two networks, and both modules are trained together jointly from scratch. 

\myparagraph{Similarity Spatial Transformer.} The final parametric layer of $\simi$ takes the spatial features produced by the ResNet backbone, flattens them, and sends them through a fully-connected layer with four output neurons $o_1$, $o_2$, $o_3$ and $o_4$. We construct the affine matrix $M$ representing the similarity transform as follows:

\begin{align}
    r &= \pi\cdot\text{tanh}(o_1) \\
    s &= \exp(o_2) \\
    t_x &= o_3 \\
    t_y &= o_4 \\
    M &= \begin{bmatrix}s \cdot \cos(r) & -s \cdot \sin(r) & t_x \\ s \cdot\sin(r) & s \cdot \cos(r) & t_y \\ 0 & 0 & 1
    \end{bmatrix}
\end{align}

To obtain the warped output image, we apply the affine matrix $M$ to an identity sampling grid, and apply the resulting transformed sampling grid to the input.

\myparagraph{Unconstrained Spatial Transformer.} The features produced from $\flow$'s ResNet backbone have spatial resolution $16\times16$. These features are fed into two separate, small convolutional networks: (1) a network that outputs a coarse flow of spatial resolution $16 \times 16$ with $2$ channels (the first channel contains the horizontal flow while the second contains the vertical flow); (2) a network that outputs weights used to perform learned convex upsampling as described in RAFT~\cite{teed2020raft}. Both of these networks consist of two conv layers (each with $3\times3$ kernels and unit padding) with a ReLU in-between. After performing $8\times$ convex upsampling on the coarse flow, we obtain our higher resolution flow $\g_{\text{dense}}$ of resolution $128 \times 128$. This flow field can be further densified with any type of interpolation.

In the case of the composed Spatial Transformer, we directly compose $\g_{\text{dense}}$ with the affine matrix $M$ predicted by $\simi$. We then sample from the original, unwarped input image according to this composed flow.

\subsection{Clustering}

For the clustering-variant of GANgealing ($K > 1$), our architectures as described above are only slightly changed as follows. $\simi$'s final fully-connected layer has $4K$ output neurons that are used to build a total of $K$ affine matrices $\{M_k\}_{k=1}^K$, one warp per cluster.

Similarly, the only change to $\flow$ is that the two convolutional networks at the end of the network each have $K$ times as many output channels---each cluster predicts its own coarse-resolution $16\times16$ flow and its own set of weights used for convex upsampling.

\myparagraph{Cluster Initialization.} In contrast to the $K = 1$ case (where we initialize $\p$ as $\bar{\w}$, the average $\w$ vector), we randomly-initialize the $\p_i$. We do this using $K$-means++~\cite{arthur2006k} (only the initialization part). In standard $K$-means++, one selects centroids from points in the input dataset. In our case, we generate a pool of 50,000 random $\w$ vectors to apply $K$-means++ to. The first centroid $\p_0$ is selected uniformly at random from the pool. Recall that $K$-means++ requires a distance function so it can gauge how well-represented the points in the pool are by the currently selected centroids. We define the distance between two latent vectors as $d(\w_1, \w_2) = \ell(G(\text{mix}(\w_1, \bar{\w})), G(\text{mix}(\w_2, \bar{\w})))$. The motivation for feeding-in $\bar{\w}$ into the later layers of StyleGAN is to ensure that the two latents are being compared based on the poses they represent when decoded to images, \textit{not} the appearances they represent. 

\subsection{Smoothly-Congealed Target Images}\label{anneal}

\begin{table}[h]\centering\vspace{-1mm}
\begin{tabular}{lcc}
 & PCK@$\alpha_\text{bbox}=0.1$\\\hline
 w/o target annealing & 59.3\%  \\
 w/ target annealing &  67.2\%  \\
\end{tabular}\vspace{1.5mm}
\caption{\textbf{The effects of smooth target annealing for LSUN Cats.} Smooth target annealing improves GANgealing performance.}
\label{anneal-abl}\vspace{-2mm}
\end{table}

A significant benefit of GAN-Supervision is that it naturally admits a smooth learning curriculum. Rather than force $T$ to learn the (often complex) mapping from $\x$ to $\y$ at the onset of training, we can instead smoothly vary the latent vector used to generate the target $\y$ from $\w$ to $\text{mix}(\p, \w)$. Early in training, $\y \approx \x$ and $T$ thus only has to learn very small warps where there are large regions of overlap between its input and target. As training proceeds, we gradually anneal the target latent $\w \rightarrow \text{mix}(\p, \w)$ with cosine annealing~\cite{loshchilov2016sgdr} over the first 150,000 gradient steps. As this occurs, the learned $\y$ target images gradually become aligned across all $\w$ and require $T$ to predict increasingly intricate warps. As shown in Table~\ref{anneal-abl}, excluding this smooth annealing degrades performance from 67.2\% to 59.3\%.

\subsection{Horizontal Flipping}\label{apdx:flip}

Because $\simi$ regresses the log-scale of a similarity transformation ($o_2$), it is incapable of performing horizontal flipping. While $\flow$ is in principle capable of learning how to horizontally (or vertically) flip an image, in practice,  we find that it is beneficial to explicitly parameterize horizontal flips. We found two methods effective.

\myparagraph{Unimodal Models.} Our unimodal models ($K=1$) are not trained with any flipping mechanism, and we introduce the capability only at test time. This is done with the flow smoothness trick as described above in Supplement~\ref{smooth}---we query $T$ with both $\x$ and $\text{flip}(\x)$, choosing whichever yields the smoothest flow field. This simple heuristic is surprisingly reliable.

\myparagraph{Clustering Models.} Clustering models provide a slightly more direct way to handle flipping. During training, when we evaluate the perceptual reconstruction loss on an input fake image $G(\w)$, we also evaluate the loss on $\text{flip}(G(\w))$ (the loss is computed against $\y = G(\text{mix}(\p, \w))$ for both inputs). We only optimize the minimum of those two losses. At test time, we need to train a classifier to predict whether an input image should be flipped. To this end, we have the cluster classifier (as described in Section~\ref{sec:clustering}) directly predict both cluster assignment as well as whether or not the input should be flipped by simply doubling the number of output classes.

\subsection{Training Details}

As with most Spatial Transformers, we initialize the final output layers of our networks such that they predict the identity transformation. For $\simi$, this is done by initializing the final fully-connected matrix as the zeros matrix (and zero bias); for $\flow$, the convolutional kernel that outputs the coarse flow is set to all-zeros. 

We jointly optimize the loss with respect to both $T$ and $\p$ using Adam~\cite{kingma2014adam} (we do not use alternating optimization). We use a learning rate of $0.001$ for $T$ and a learning rate of $0.01$ for $\p$. We apply the cosine annealing with warm restarts scheduler~\cite{loshchilov2016sgdr} to both learning rates. We use a batch size of 40. Finally, we note that we do not make use of any data augmentation---GANgealing uses raw samples directly from the generator as the training data.

\subsection{Hyperparameters}

\begin{table}
    \vspace{3mm}
    \hfill
    \resizebox{\linewidth}{!}{
    \begin{tabular}{ lccccc }
    \toprule 
    \textbf{Dataset} & \textbf{$\mathcal{W}^+$ cutoff} & $N$ & $K$ & \textbf{GAN Resolution} \\ \hline
    LSUN (unimodal) & 5 & 1 & 1 & 256$\times$256 \\
    CUB & 5 & 1 & 1 & 256$\times$256 \\
    LSUN (clustering) & 6 & 5 & 4 & 256$\times$256 \\
    In-The-Wild CelebA & 6 & 512 & 1 & 128$\times$128 \\
    \bottomrule
    \end{tabular}
    }
    \vspace{-2mm}
    \caption{\textbf{GANgealing Hyperparameters}.\vspace{-3mm}}\label{hyperparams}

\end{table}

We use $\lambda_{TV} = 2500$ for all models trained with LPIPS and $\lambda_{TV} = 1000$ for all models trained with the self-supervised perceptual loss. All models use $\lambda_I = 1$. We detail other hyperparameters in Table~\ref{hyperparams}.

$N$ controls how many degrees of freedom $\p$ has in choosing the target pose that $T$ congeals towards. As shown in Table~\ref{pck-abl}, small $N$ are \textit{critical} for some models (LSUN Cats with $N=512$ gets less than 1\% PCK on SPair Cats whereas $N=1$ achieves 67\%). However, StyleGAN2 generators with less expressive $\mathcal{W}$ spaces (such as those trained on In-The-Wild CelebA) can function well without any constraints ($N=512$). 

\myparagraph{Padding Mode.} One subtle hyperparameter is the padding mode of the Spatial Transformer which controls how $T$ samples pixels beyond image boundaries. We found that $\texttt{reflection}$ padding is the ``safest" option and seems to work well in general. We also found $\texttt{border}$ padding works well on some datasets (e.g., LSUN Cats and Dogs, CUB, In-The-Wild CelebA), but can be more prone to degenerate solutions on datasets like LSUN Bicycles and TVs. We recommend $\texttt{reflection}$ as a default choice.

\begin{figure}
    \centering
    \includegraphics[width=\linewidth]{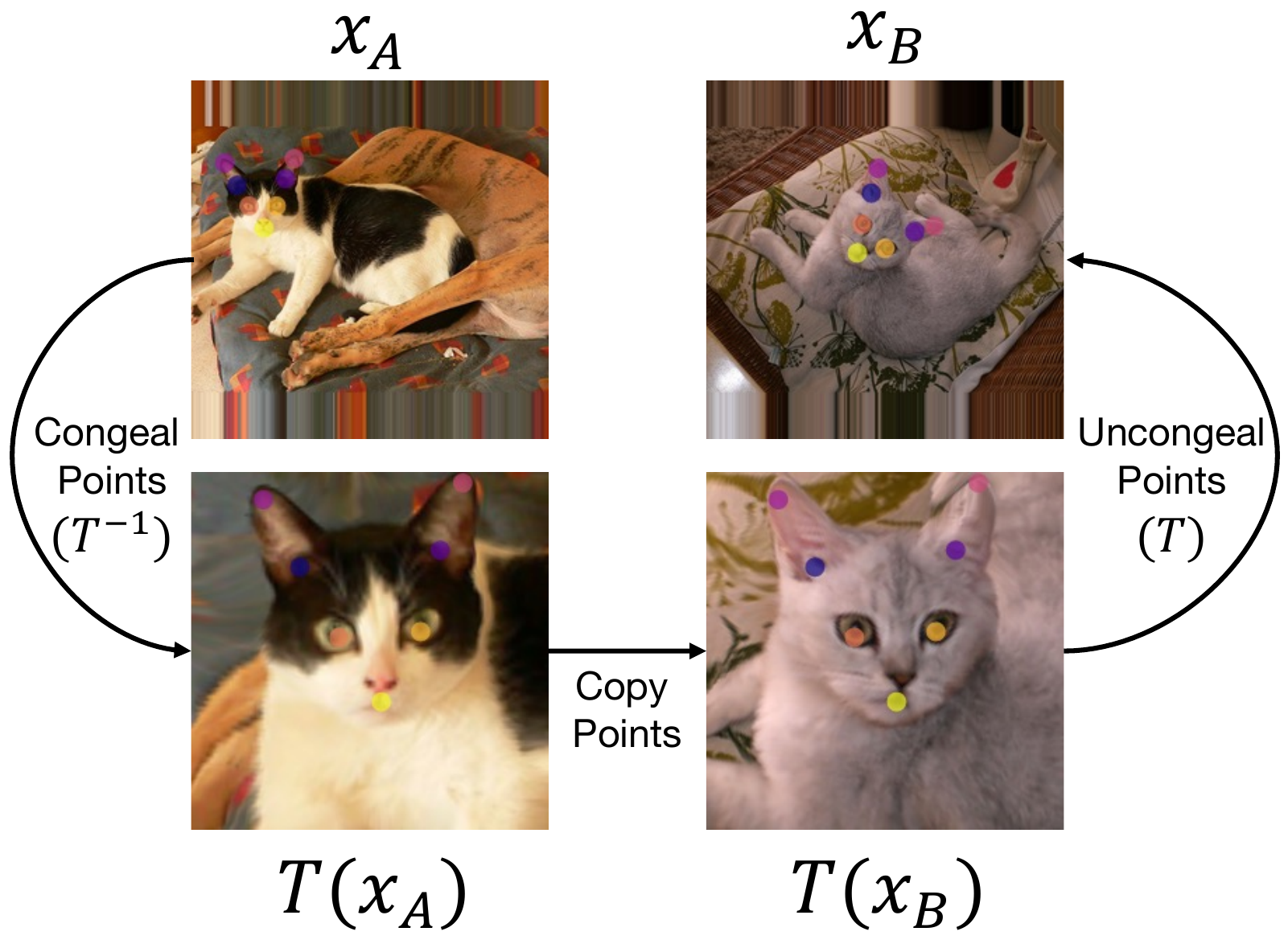}
    \caption{\textbf{The process for finding correspondences between two images with GANgealing}.}
    \label{transfer_method}
\end{figure}

\subsection{Direct Image-to-Image Correspondence}\label{apdx:transfer}

Our method is able to find dense correspondences directly between any pair of images $\x_A$ and $\x_B$.  Figure~\ref{transfer_method} gives an overview of the process. At a high-level, this merely involves applying the forward warp that maps points in $\x_A$ to points in $T(\x_A)$ and composing it with the reverse warp that maps points in the congealed coordinate space back to $\x_B$. We go into detail about this procedure in this section.

Recall that $T$ outputs a sampling grid $\g$ which maps points in congealed space to points in the input image. Without loss of generality, say we wish to know where a point $(i, j)$ in $\x_A$ corresponds to in $\x_B$. First, we need to determine where $(i, j)$ maps to in the congealed image $T(\x_A)$---i.e., we need to \textit{congeal} point $(i, j)$. This is given by the value at pixel coordinate $(i, j)$ in the \textbf{inverse} of the sampling grid produced by $T$ for $\x_A$. When $T$ produces similarity transformations, we can analytically compute this inverse by inverting the affine matrix representing the similarity transform and applying it to the identity sampling grid. Unfortunately, for the unconstrained flow case we cannot analytically determine $\g^{-1}$. There are many ways one could go about obtaining this inverse. We opt for the simplest solution---inversion via nearest neighbors. Specifically, we can approximate the quantity by using nearest neighbors to find the pixel coordinates that give rise to the coordinates closest to $(i, j)$ in $\g$. Recall that $\g_{i, j}$ is the input pixel coordinate that gets mapped to pixel $(i, j)$ in congealed coordinate space. Then we can approximate $\g^{-1}_{i, j} \approx \argmin_{i', j'} ||\g_{i', j'} - [i,j]||_2$. 

Now that we know where points in $\x_A$ map to in $T(\x_A)$, the last step is to determine where the congealed points ($\g^{-1}_{i, j}$) map to in $\x_B$---i.e., we need to \textit{uncongeal} $\g^{-1}_{i, j}$. This is the easy step: we need only query the sampling grid (produced by applying $T$ to $\x_B$) at location $\g^{-1}_{i, j}$.

\begin{figure*}
    \centering
    \includegraphics[width=\linewidth]{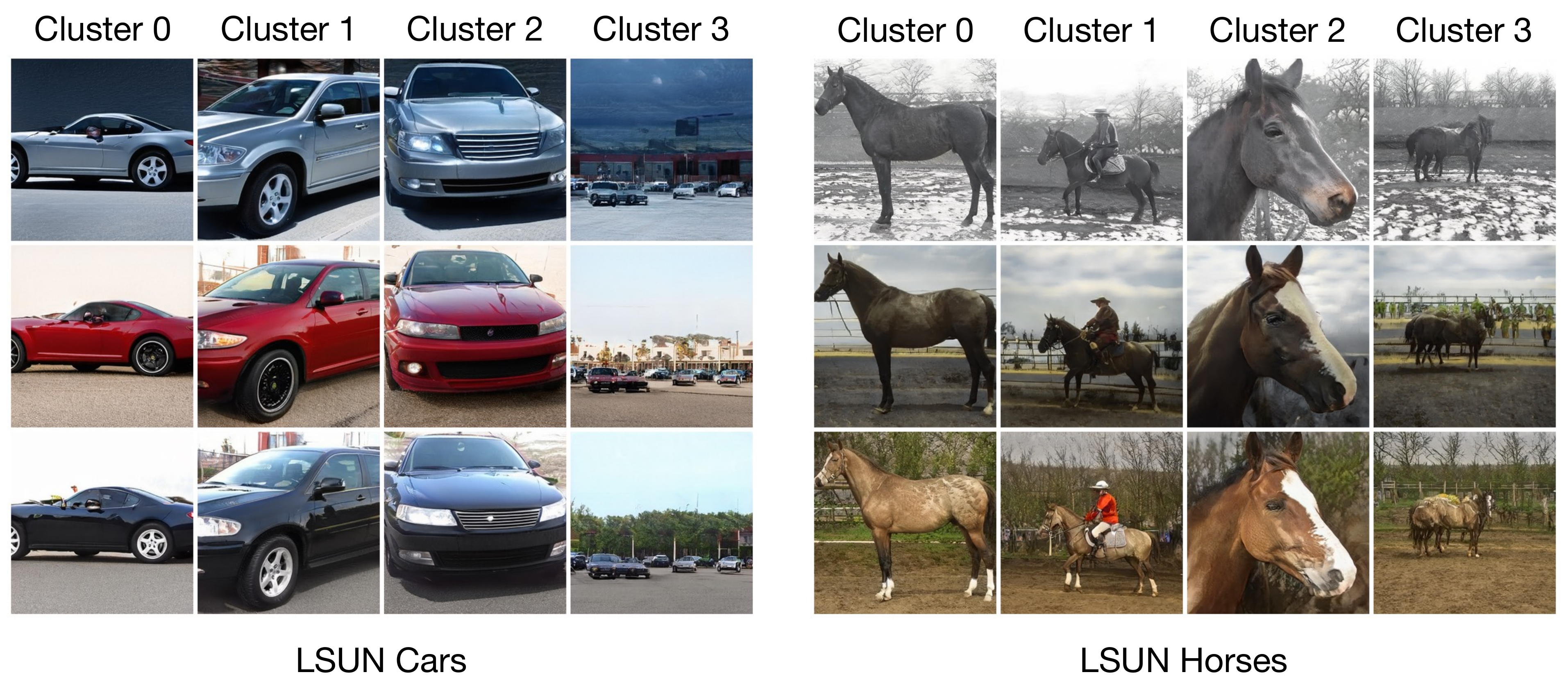}
    \caption{\textbf{Learned clusters}. We show three samples of target images $\y$ at the end of training which define our $K=4$ learned clusters.}
    \label{fig:cluster_modes}
\end{figure*}

\begin{figure*}
    \centering
    \includegraphics[width=\linewidth]{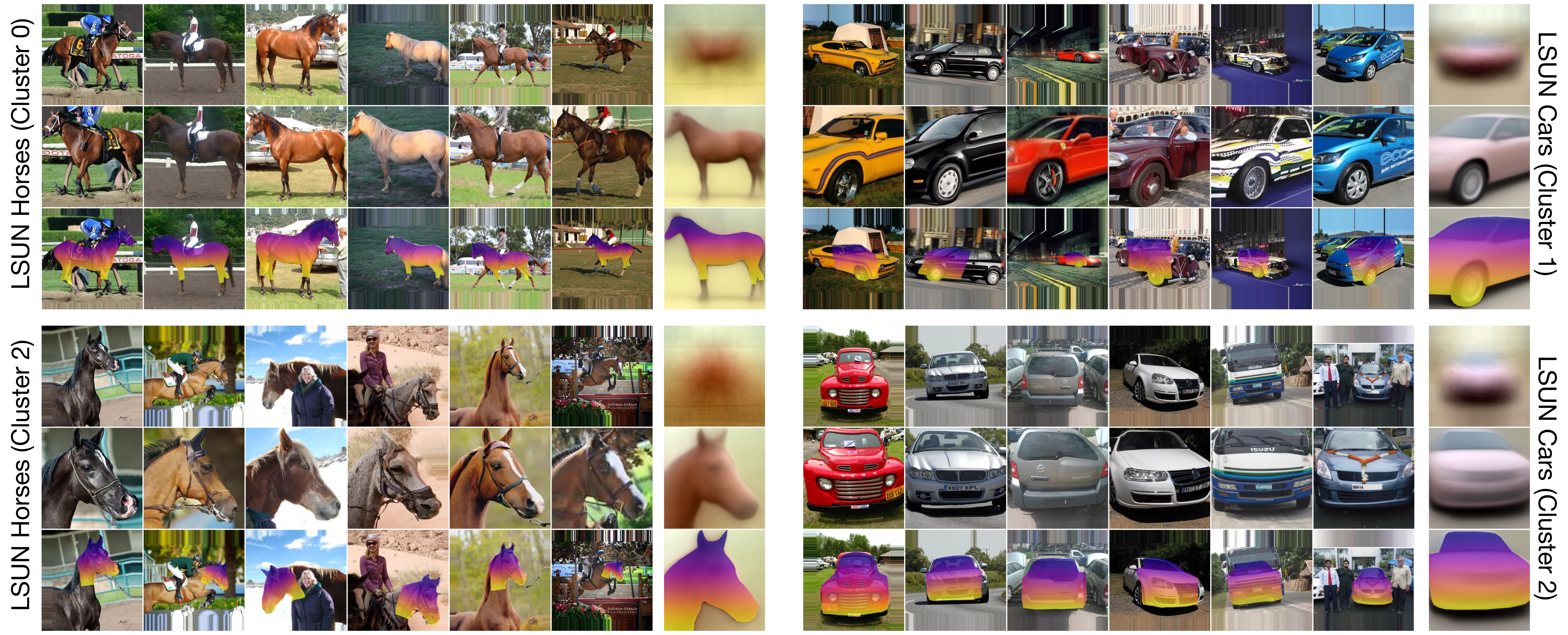}
    \caption{\textbf{Dense correspondence results for different clusters}. For each cluster, the top row shows unaligned real images assigned to that cluster by our cluster classifier; we also show the cluster's average 
    image. The middle row shows our learned transformations of the input images. The bottom row shows dense correspondences between the images.}
    \label{fig:cluster_variety}
\end{figure*}

\section{Clustering Results}

Figure~\ref{fig:cluster_modes} shows our $K=4$ learned clusters for our LSUN Cars and Horses models. We show dense correspondence results for various clusters in Figure~\ref{fig:cluster_variety}.

\section{Visualizing GAN-Supervised Training Data}

\begin{figure*}
    \centering
    \includegraphics[width=\linewidth]{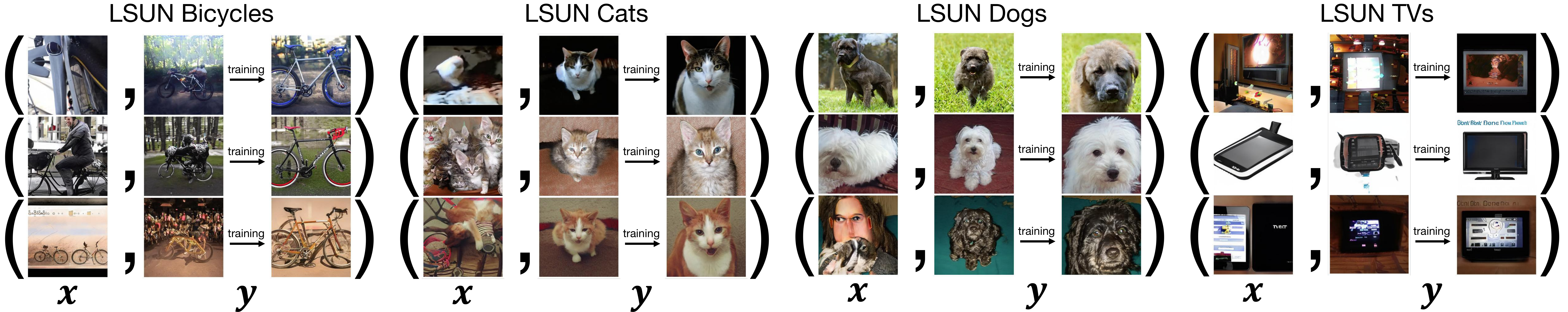}
    \caption{\textbf{Examples of GAN-Supervised paired data used in GANgealing}. For each dataset, we show random GAN samples $\x=G(\w)$ used to train our Spatial Transformer. For each input $\x$, we show both the \textit{initial} target image at the start of training $\y = G(\text{mix}(\p=\bar{\w}, \w))$ as well as our \textit{learned} target at the end of training $\y = G(\text{mix}(\p, \w))$. The initial target images (initialized with the truncation trick) are often unrealistic---for example, the initial LSUN TV targets are largely incoherent. Learning $\p$ is essential so images can be congealed to a coherent target. Note that we omit target annealing (Supplement~\ref{anneal}) in this visualization for clarity.}
    \label{fig:gan_supervision}
\end{figure*}

We show examples of our paired GAN-Supervised training data in Figure~\ref{fig:gan_supervision}. In Table~\ref{pck-abl}, we showed that learning the fixed mode can be essential for some datasets (e.g., LSUN Cats). Figure~\ref{fig:gan_supervision} illustrates one potential reason why it is so critical: often, the initial truncated target mode in StyleGAN2 models produces \textit{unrealistic} images. For example, truncated bicycles are surrounded with incoherent texture and have an unnatural structure, truncated TVs are unintelligible and truncated cats have unnatural bodies. Congealing all images to these poor targets could produce erroneous correspondences; hence, \textit{learning} the target mode is in general very important.

Furthermore, the initial mode is often unsuitable as a dataset-wide congealing target. For example, not all LSUN Cat images (real or fake) feature the full-body of a cat, but most do feature a cat's upper-body. Hence, GANgealing updates the target to a more suitable target ``reachable" by the broader distribution. Finally, also observe that our Spatial Transformer can be successfully trained even in the presence of \textit{imperfect targets}: the target bicycles sometimes do not retain the color of the corresponding input fake bicycle, for instance.  

\begin{figure}
    \centering
    \includegraphics[width=\linewidth]{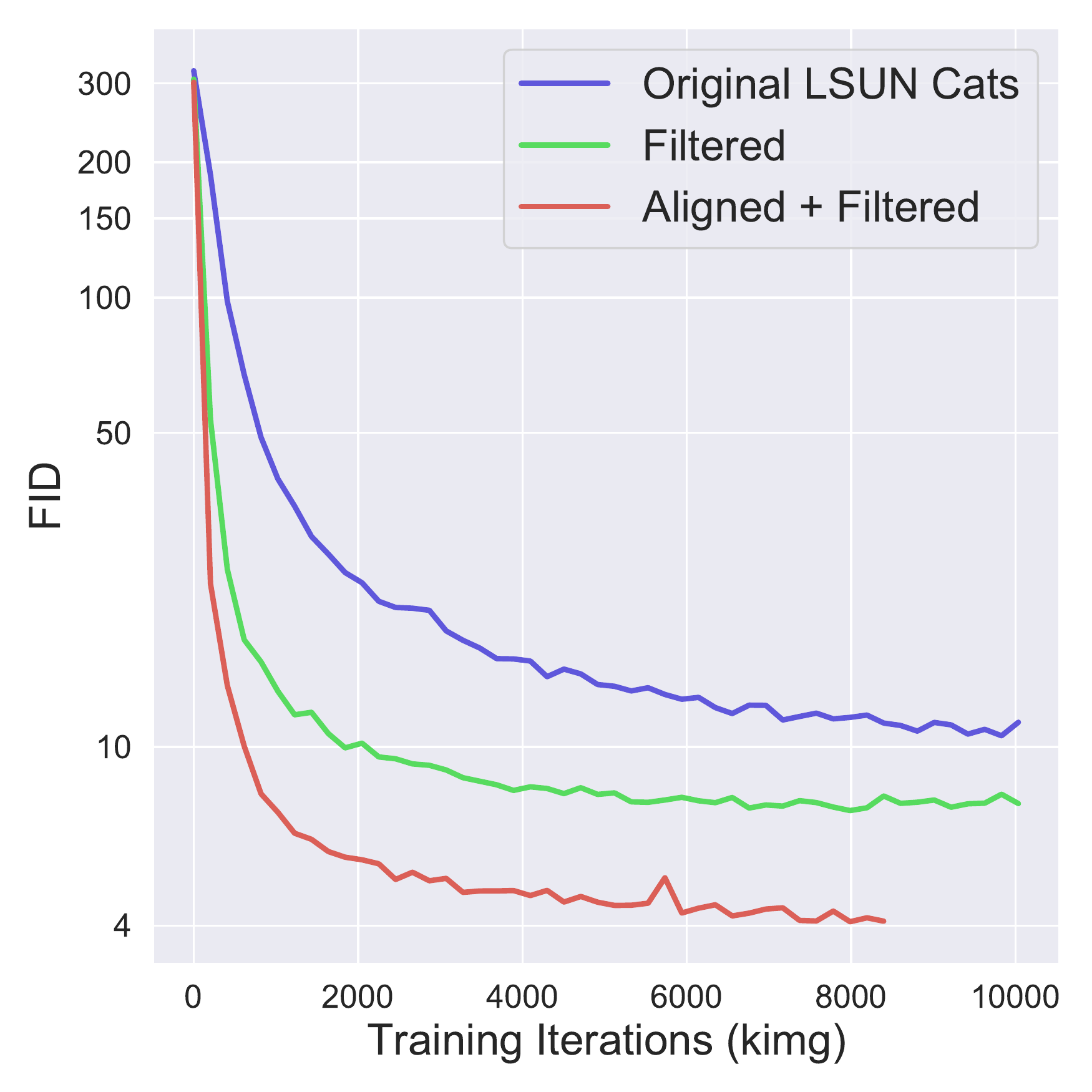}
    \caption{\textbf{The effect of aligning and filtering datasets before GAN training}. Each curve represents a StyleGAN2 trained on LSUN Cats with different data pre-processing. For each model, we compute FID against its corresponding pre-processed real distribution (i.e., the Original LSUN Cats curve computes FID against LSUN Cats, Filtered LSUN Cats computes FID against the filtered LSUN Cats distribution, etc.). Aligning and filtering images with our method yields significantly better FID early in training by simplifying the real distribution.}
    \label{fig:fid50k_full}
\end{figure}

\section{Accelerating GAN Training with Learned Pre-Processing}\label{accel}

A natural application of GANgealing is unsupervised dataset alignment for downstream machine learning tasks. In this section, we use our trained Spatial Transformer networks to align and filter data for GAN training. 

\myparagraph{Filtering.} The first step is dataset filtering. We would like to remove two types of images: (1) images for which our Spatial Transformer makes a mistake (i.e., produces an erroneous alignment) and (2) images that are \textit{unalignable}. For example, several images in LSUN Cats do not actually contain any cats. And, some images contain cats that cannot be well-aligned to the target mode (e.g., due to significant out-of-plane rotation of the cat's face). We can automatically detect these problematic images by examining the \textit{smoothness} of the flow field produced by the Spatial Transformer for a given input image: images with highly un-smooth flows usually correspond to one of these types of problematic images. Please refer to Supplement~\ref{smooth} below for a detailed explanation of this procedure. Here, we experiment with dropping 75\% of real images (those with the least-smooth flow fields), reducing LSUN Cats' size from 1,657,264 images to 414,316 images.

\myparagraph{Alignment.} The second step is to align the filtered dataset. This is done by applying our Spatial Transformer to congeal every image in the input dataset. To avoid introducing excessive warping artifacts into the output distribution, we only use our similarity Spatial Transformer $\simi$ (which performs oriented cropping) in this step---we merely remove the unconstrained $\flow$ Spatial Transformer module from our trained $T$. To ensure a high quality output dataset, we remove some additional images during this procedure. (1) We removes images for which $\simi$ has to extrapolate a large number of pixels beyond image boundaries to avoid output images with lots of visible padding. (2) We remove images where $\simi$ zooms-in ``too much." The motivation for this second criterion is that some images contain very small objects, and in these cases the congealed image will be low resolution (blurry). These two heuristics further reduce dataset size from 414,316 images to 58,814 high quality aligned images.

\myparagraph{Quantitative Results.} We apply our learned pre-processing procedure to the LSUN Cats dataset for downstream GAN training. As is common practice when training GANs on aligned datasets like AFHQ and FFHQ, we use mirror augmentations~\cite{karras2019style} when training on our aligned data. In Figure~\ref{fig:fid50k_full}, we show that training with our learned pre-processing enables GANs to converge to good FID~\cite{heusel2017gans} faster by simplifying the training distribution. We also show that while \textit{only} filtering the dataset to 58,814 images without alignment accelerates convergence (a conclusion similar to the one found by DeVries et. al~\cite{devries2020instance}), both steps together provide the greatest speed-up. We stress that our gains in Figure~\ref{fig:fid50k_full} come from \textit{simplifying the training distribution}. We leave showing how learned alignment can improve performance on the original, unaligned distribution to future work. Given the extent to which human-supervised dataset alignment and filtering is prevalent in the GAN literature~\cite{choi2020starganv2,karras2018progressive,karras2019style}, we hope our unsupervised procedure will be used in the future to automate these important steps.

\begin{figure*}[h!]
    \centering
    \includegraphics[width=0.85\linewidth]{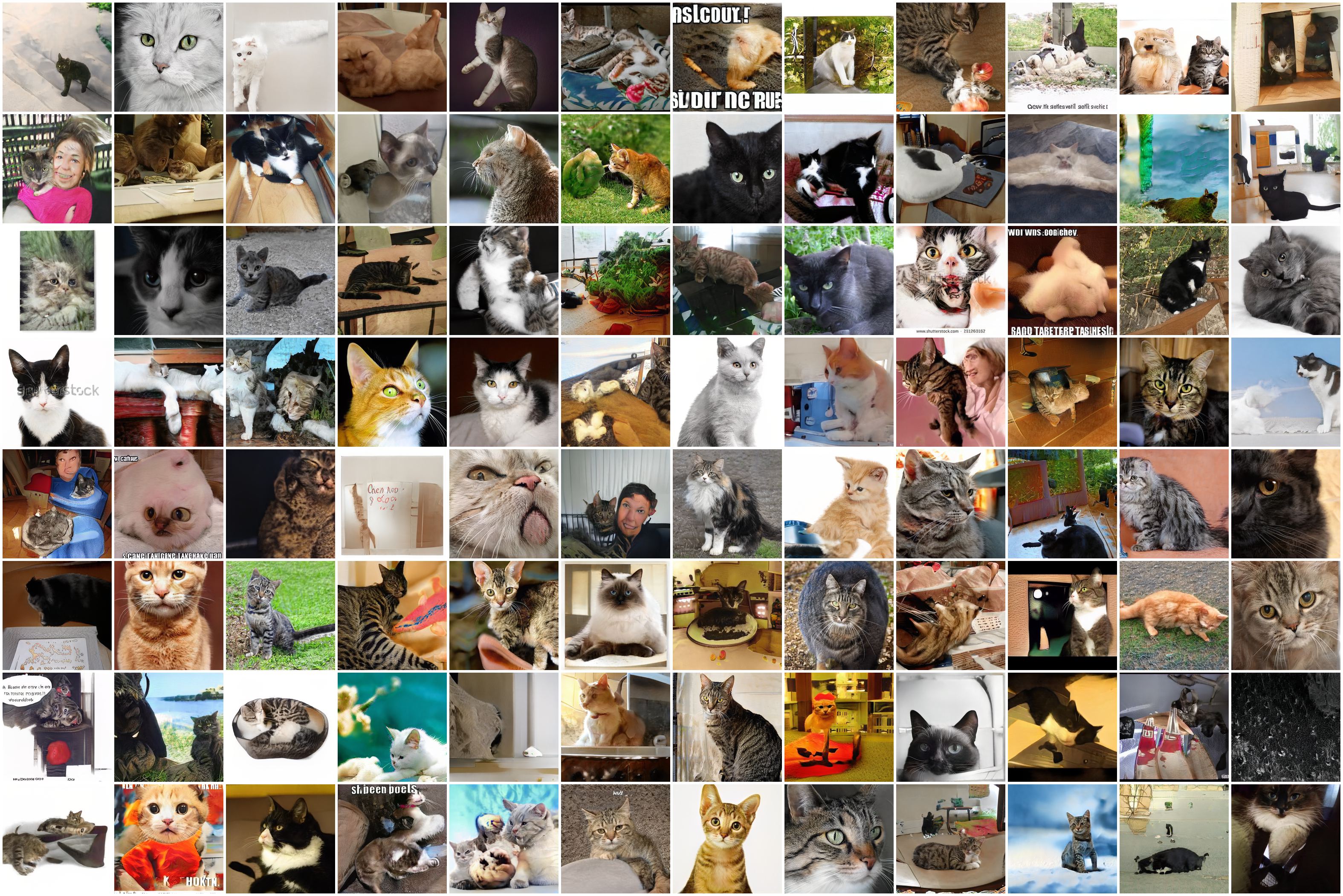}
    \caption{\textbf{Uncurated samples from an LSUN Cats StyleGAN2}.}
    \label{sg2}
\end{figure*}

\begin{figure*}[h!]
    \centering
    \includegraphics[width=0.85\linewidth]{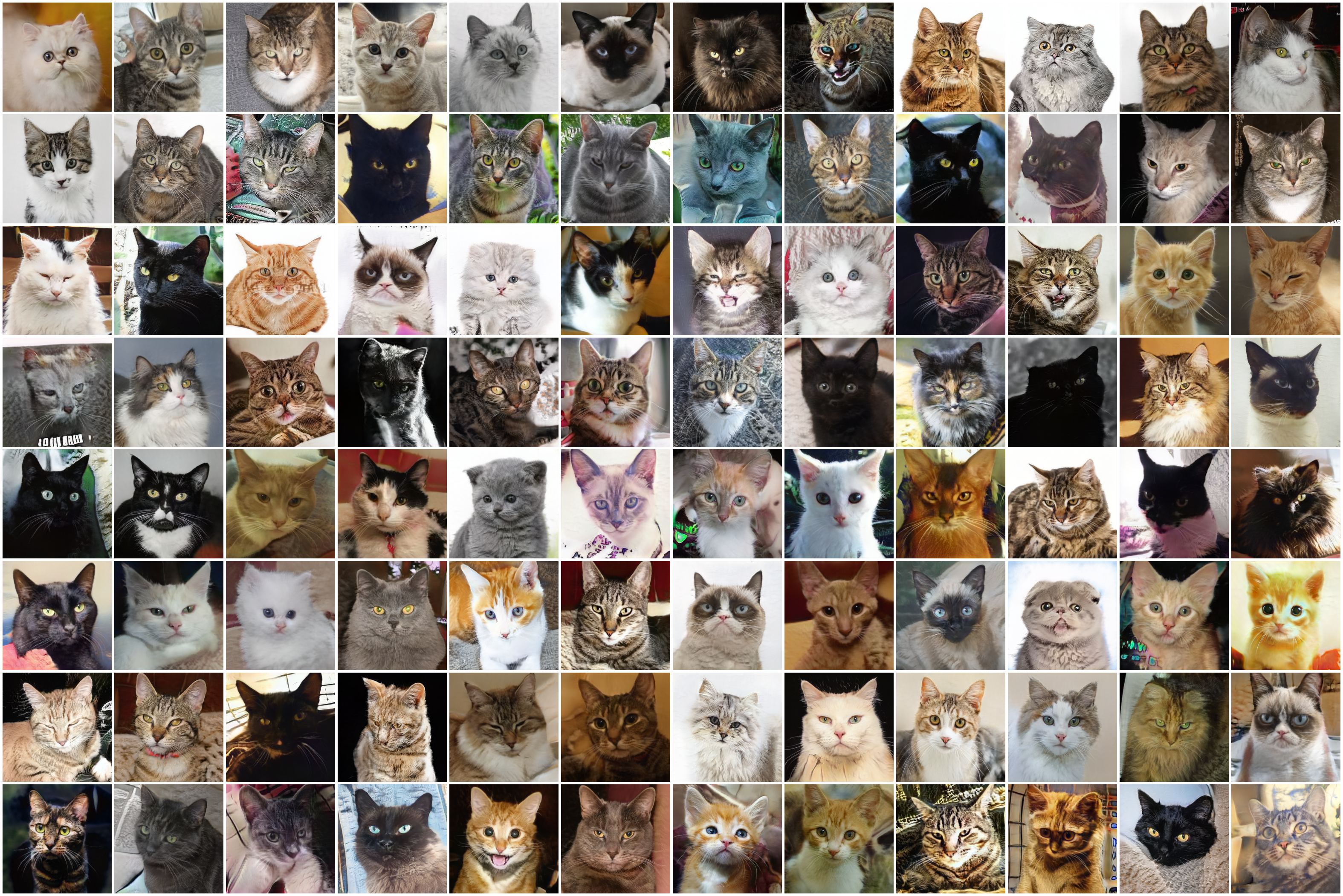}
    \caption{\textbf{Uncurated samples from a StyleGAN2 trained on LSUN Cats pre-processed with our learned alignment and filtering}. Our method leads to a GAN with higher overall visual fidelity at the cost of reduced dataset complexity.}
    \label{aligned}
\end{figure*}

\myparagraph{Visual Results.} In Figure~\ref{sg2} we show 96 uncurated, untruncated GAN samples for the baseline LSUN Cats model (FID versus LSUN Cats is 6.9); in Figure~\ref{aligned} we show uncurated samples from a StyleGAN2 trained on our learned aligned and filtered LSUN Cats (FID versus pre-processed LSUN Cats is 3.9). Our learned alignment improves the visual fidelity of the generator by reducing the complexity of the real distribution through alignment and filtering.

\section{Uncurated Dense Correspondence Results}\label{apdx:uncurated}

Below we show 120 \textit{uncurated} dense correspondence results for each of our eight datasets. The results are best viewed zoomed-in.

\begin{figure}[!t]
    \centering
    \includegraphics[width=0.97\linewidth]{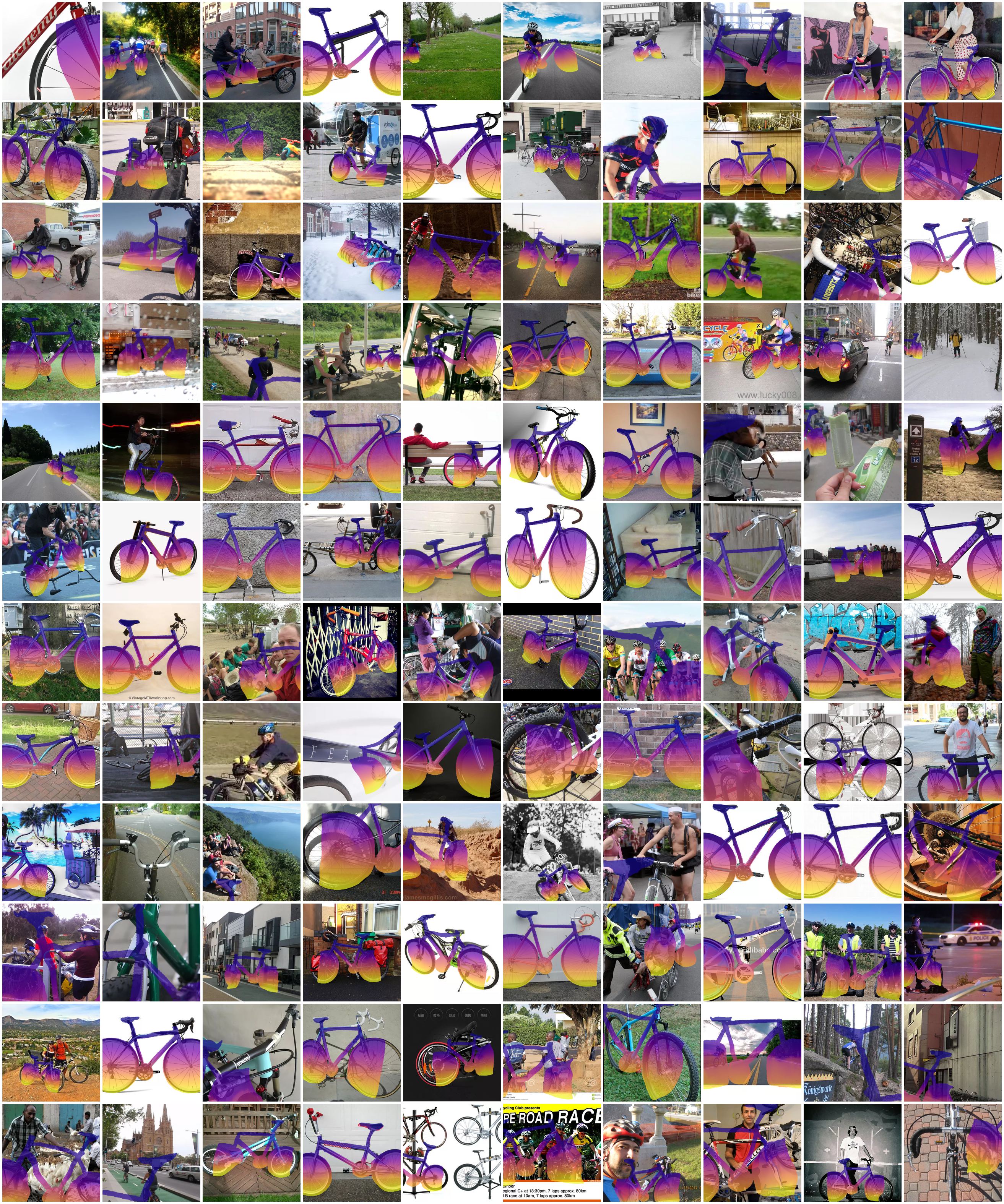}
    \caption{\textbf{LSUN Bicycles uncurated results}.}
    \label{fig:bicycle}
\end{figure}

\begin{figure}[!t]
    \centering
    \includegraphics[width=0.97\linewidth]{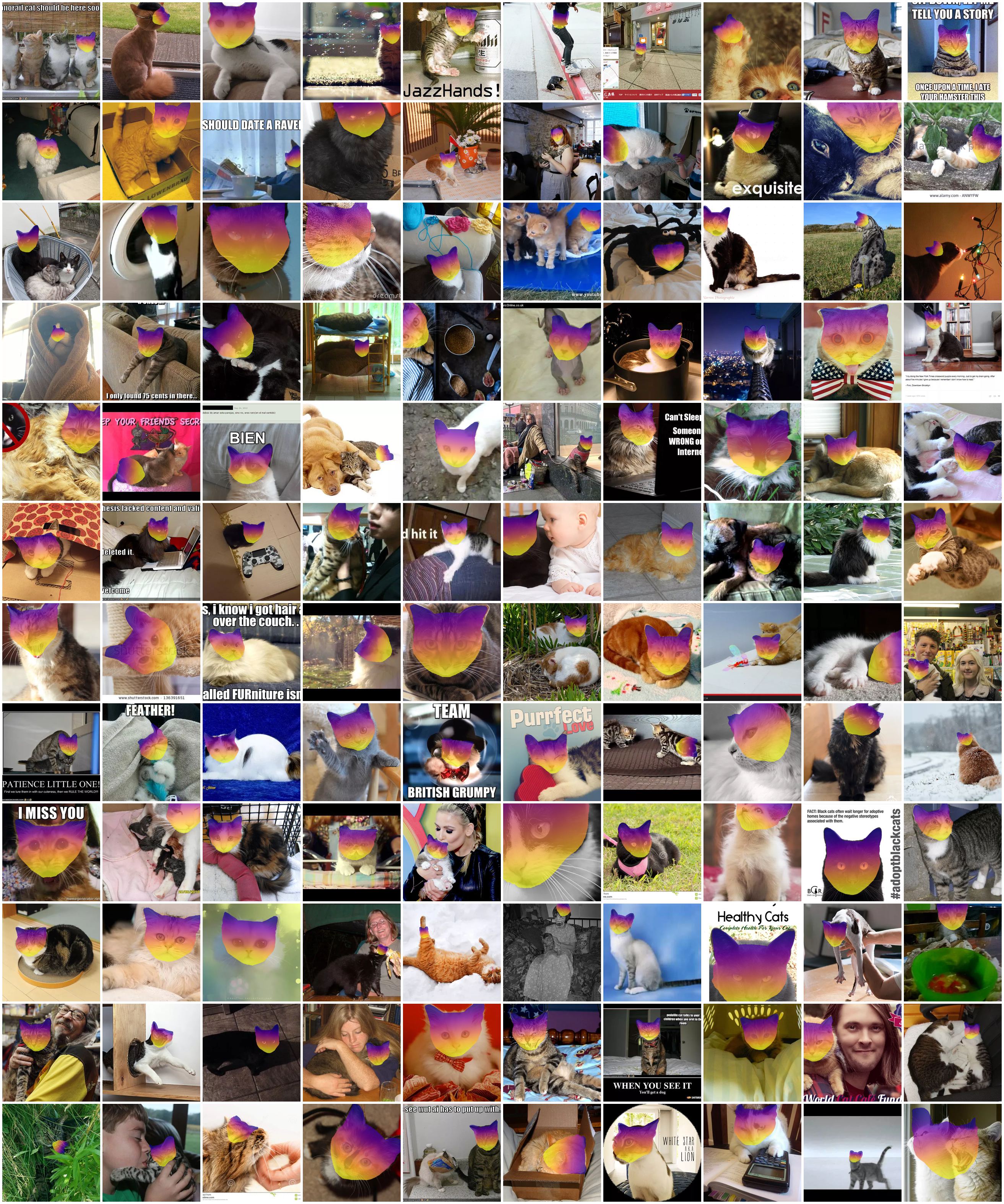}
    \caption{\textbf{LSUN Cats uncurated results}.}
    \label{fig:cat}
\end{figure}

\begin{figure}[!t]
    \centering
    \includegraphics[width=0.97\linewidth]{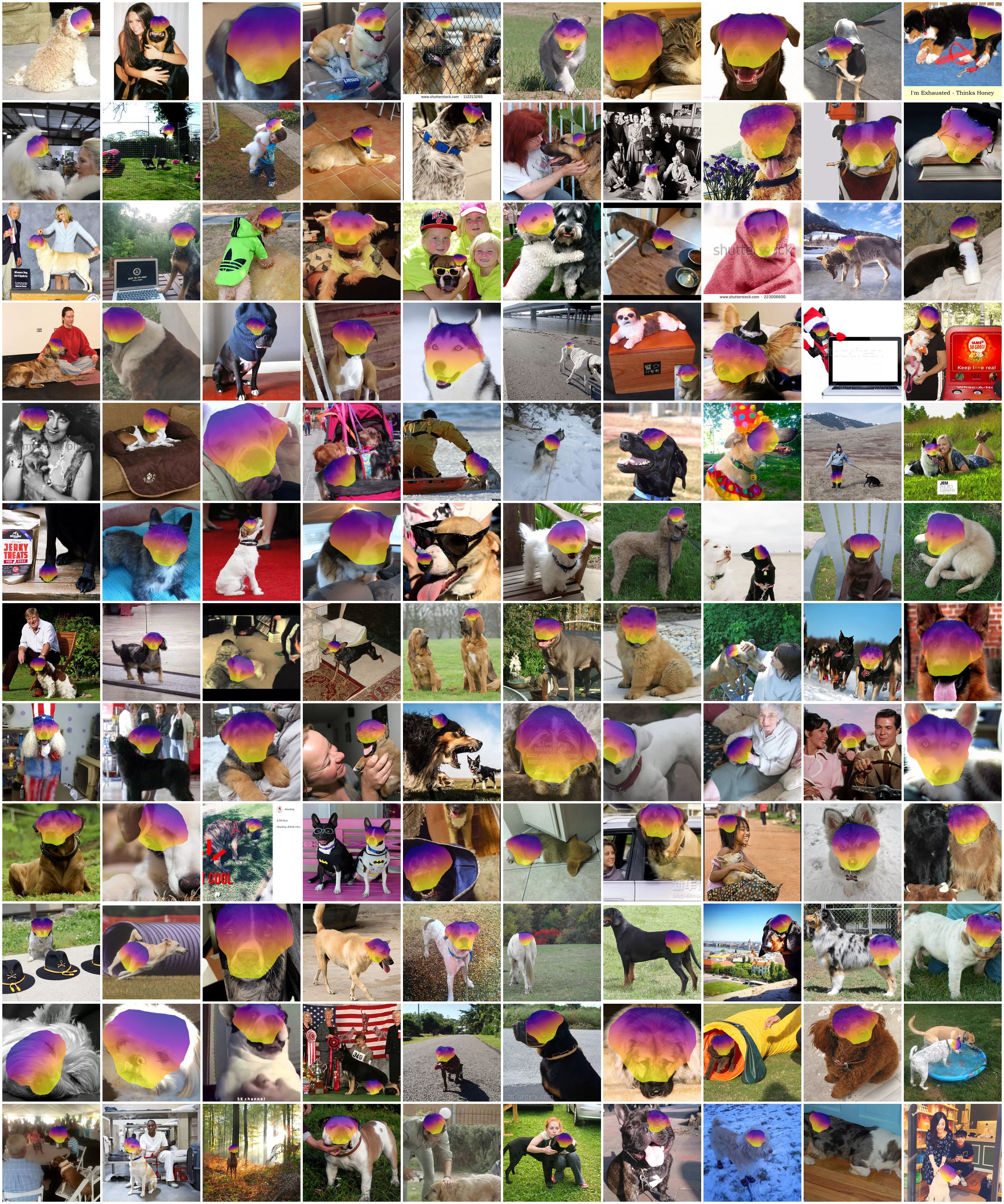}
    \caption{\textbf{LSUN Dogs uncurated results}.}
    \label{fig:dog}
\end{figure}

\begin{figure}[!t]
    \centering
    \includegraphics[width=0.97\linewidth]{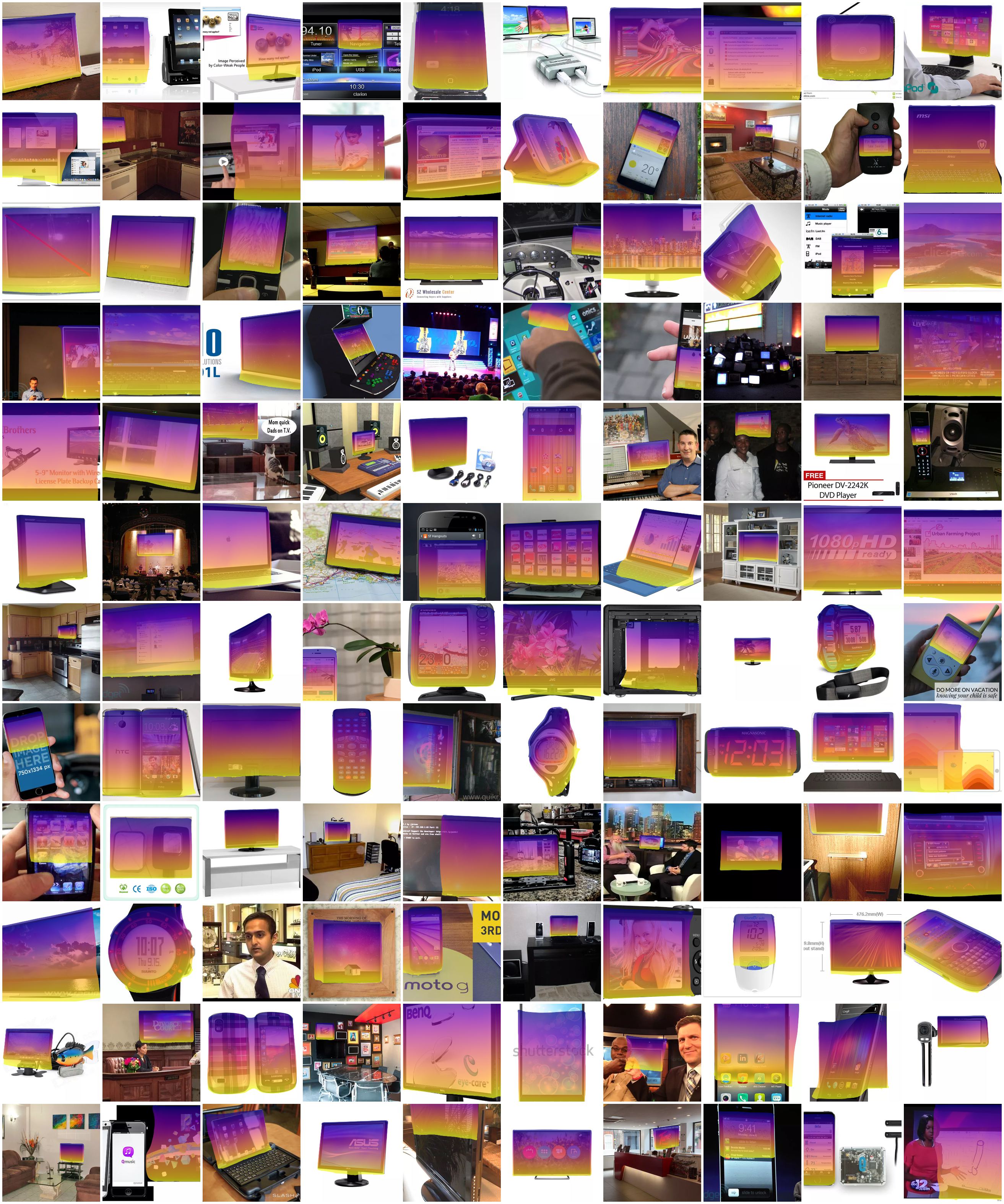}
    \caption{\textbf{LSUN TVs uncurated results}.}
    \label{fig:tv}
\end{figure}

\begin{figure}[!t]
    \centering
    \includegraphics[width=0.97\linewidth]{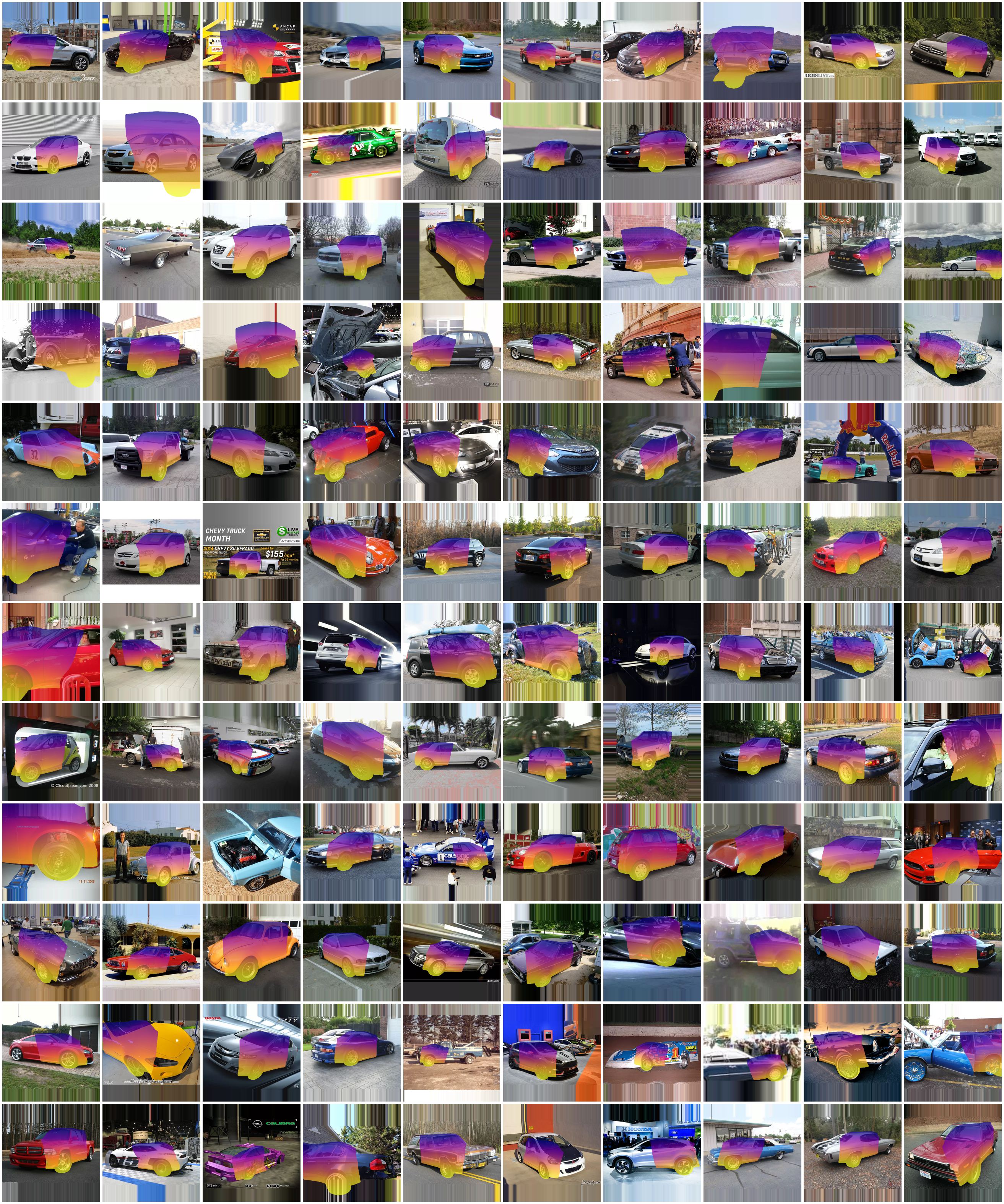}
    \caption{\textbf{LSUN Cars (cluster 1) uncurated results}.}
    \label{car}
\end{figure}

\begin{figure}[!t]
    \centering
    \includegraphics[width=0.97\linewidth]{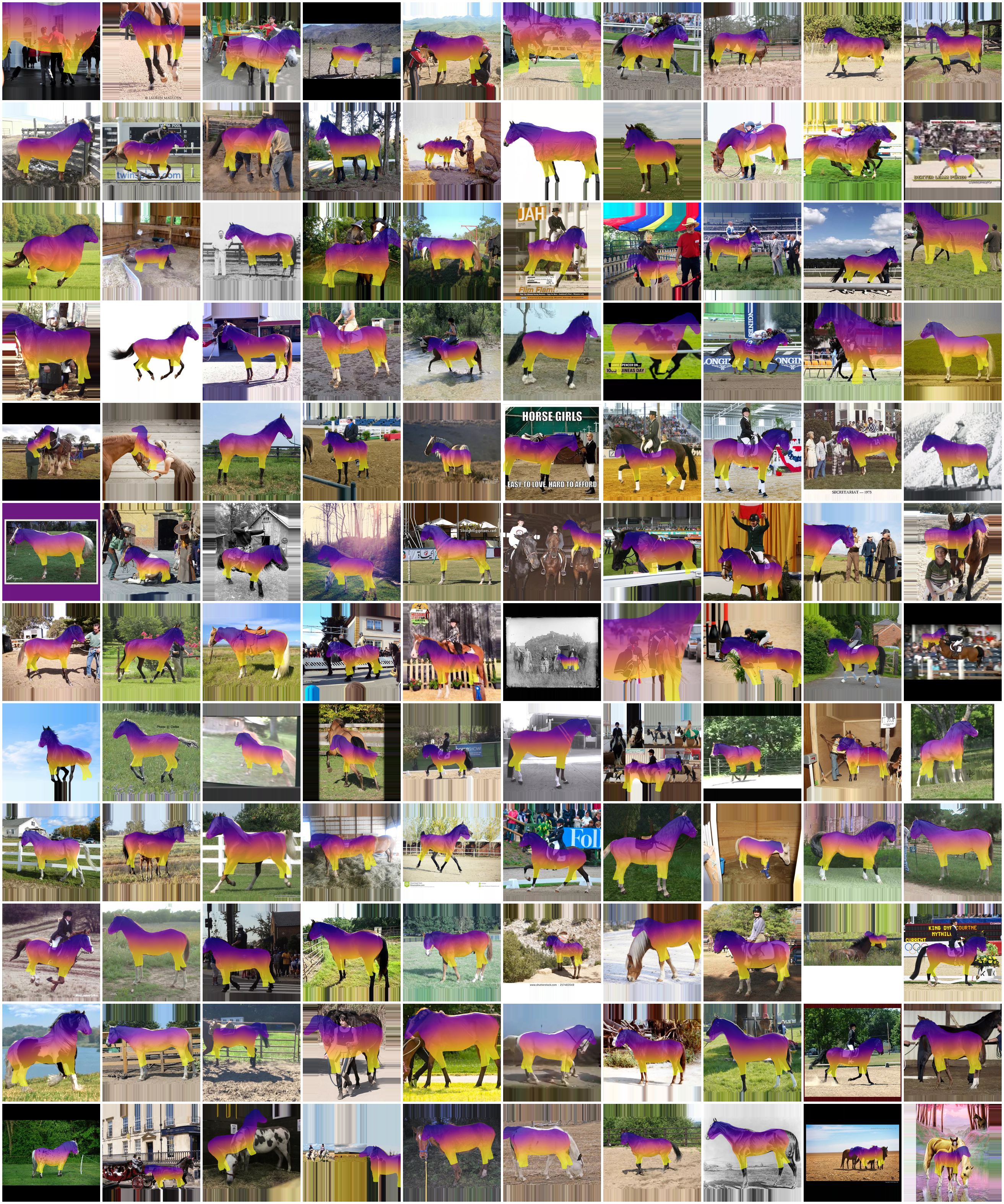}
    \caption{\textbf{LSUN Horses (cluster 0) uncurated results}.}
    \label{horse}
\end{figure}

\begin{figure}[!t]
    \centering
    \includegraphics[width=0.97\linewidth]{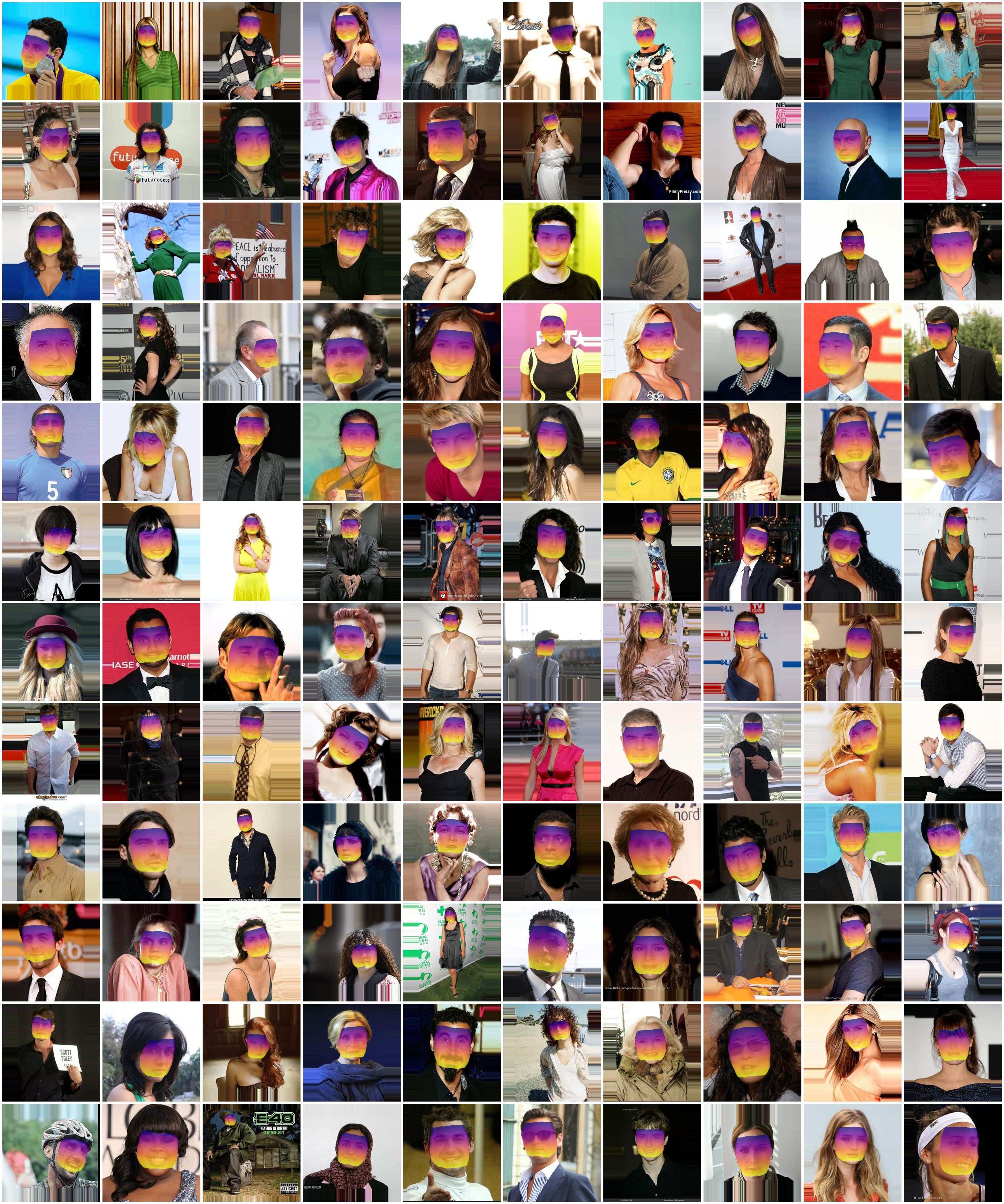}
    \caption{\textbf{In-The-Wild CelebA results}. The results are uncurated with the exception of replacing three potentially offensive images.}
    \label{celeba}
\end{figure}

\begin{figure}[!t]
    \centering
    \includegraphics[width=0.97\linewidth]{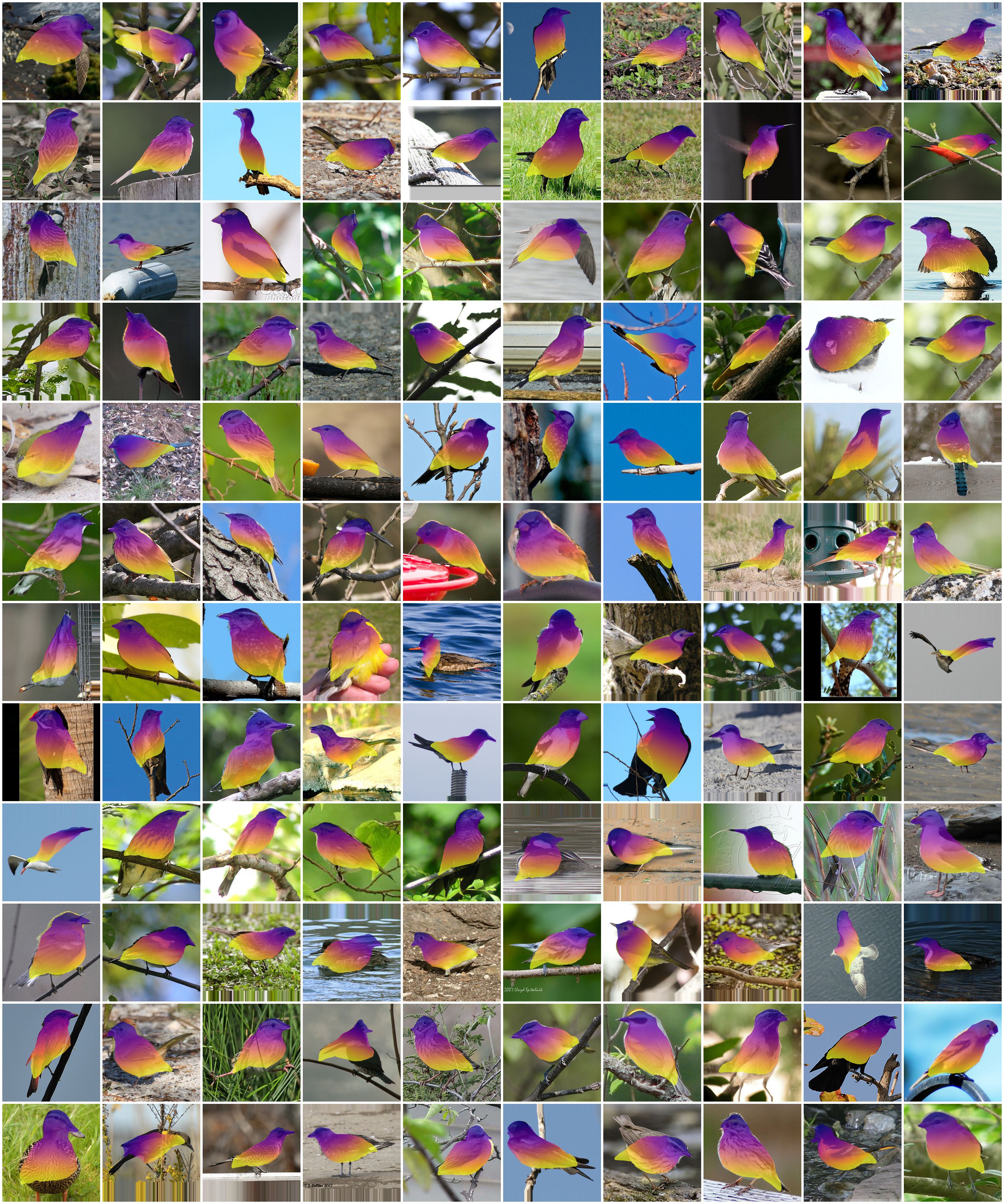}
    \caption{\textbf{CUB uncurated results}.}
    \label{cub}
\end{figure}

\section{Emergent Properties of GANgealing}

We empirically find that our Spatial Transformer develops two useful test time capabilities from training with GANgealing.

\subsection{Recursive Alignment}

Recall that our Spatial Transformer consists of two submodules: a similarity warping Spatial Transformer $\simi$ and an unconstrained Spatial Transformer $\flow$. Once trained, we find that $\simi$'s primary role is to localize objects and correct in-plane rotation; $\flow$ brings the localized object into tight alignment by handling articulations, out-of-plane rotation, etc. For especially complex datasets (e.g., all LSUN categories), objects appear at many scales and are sometimes non-trivial to accurately localize. Surprisingly, we find that the $\simi$ network can be applied \textit{recursively} multiple times to its own output at test time, significantly improving the ability of $T$ to accurately align challenging images. This recursive processing appears to be relatively stable---i.e., $T$ does not explode as we increase the number of recursions. We use three recursive iterations of $\simi$ for our LSUN models.

We suspect the reason this behavior emerges is because, over the course of training, it is likely that many generated inputs $\x$ will be sampled that are close to the target $\y$. In other words, it is likely that $\w$ get sampled such that $\w \approx \p$\interfootnotelinepenalty=10000\footnote{This is likely \textit{as long as $\p$ is properly constrained}. When poorly constrained, $\p$ may correspond to an \textit{out-of-distribution} image and hence $T$ may not be used to processing images similar to sampled $\y$. Indeed, for our In-The-Wild CelebA model where $N=512$ (i.e., $\p$ is not constrained at all), we find recursion is unstable. For all LSUN models where $N\leq 5$, we find it helps significantly.}, and hence the Spatial Transformer must learn to produce approximately the identity function in these cases. Thus, aligned fake images happen to be \textit{in-distribution} for $T$, meaning it can stably process increasingly aligned real images at test time without issue in this recursive evaluation.

\begin{figure}
    \centering
    \includegraphics[width=\linewidth]{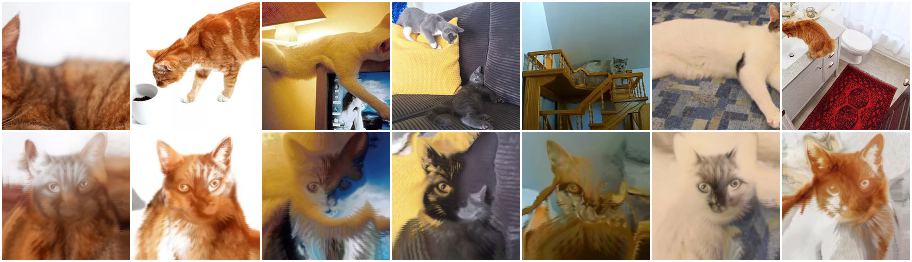}
    \caption{\textbf{GANgealing produces high frequency flow fields for unalignable images}. Each image in the top row shows a cat that either cannot be congealed to the learned mode or is a failure case of the Spatial Transformer. Our $T$ produces high frequency flows to try and ``fool" the perceptual loss for these hard cases. These examples are easily detected.}
    \label{high_freqs}
\end{figure}

\subsection{Flow Smoothness}\label{smooth}

A particularly useful behavior developed by our $T$ is a tendency to \textit{fail loudly}. What happens when $T$ is faced with an unalignable image or makes a mistake? As shown in Figure~\ref{high_freqs}, $T$ produces very high-frequency flow fields to try and fool the perceptual loss (e.g., by synthesizing cat ears and eyes from background texture). It turns out these failure cases are easily detectable by simply measuring how smooth the flow field produced by $T$ is---this can be done by evaluating our $\mathcal{L}_{TV}$ total variation loss on the flow. This behavior enables several test time capabilities; for example, we can determine if an image should be horizontally flipped by merely running $\x$ and $\text{flip}(\x)$ through $T$ and selecting whichever input yields the smoothest flow. Or, we can score real images by their flows' $\mathcal{L}_{TV}$ and drop a percentage of the dataset with the worst (most positive) scores---this is how we perform dataset filtering with our $T$ as part of our learned pre-processing for downstream GAN training as described above in Supplement~\ref{accel}.

\section{Performance}

\myparagraph{Training.} Our unimodal GANgealing models train for 1.3125M gradient steps; this takes roughly 48 hours on 8$\times$ A100 GPUs for 256$\times$256 StyleGAN2 models. Over the course of training, our Spatial Transformers process more than 50M randomly-sampled GAN images.

\myparagraph{Inference.} In this section, we briefly discuss runtime comparisons between GANgealing and supervised baselines. We compare the time to perform the cartoon cat face augmented reality application online (batch size of 1, RTX 6000 GPU). This involves propagating over 3.9 million points every frame. GANgealing runs at 15 FPS, CATs~\cite{cho2021semantic} at 13 FPS, and RAFT~\cite{teed2020raft} at 3 FPS. For time to perform keypoint transfers between pairs of images (SPair Cats), GANgealing runs at 31 pairs per second, and CATs runs at 70 pairs per second. We note that we did not optimize our implementation for inference, and so we expect it is possible to further improve performance.

\section{Broader Impacts}

As with all machine learning models, our Spatial Transformer is only as good as the data on which it is trained. In the case of GANgealing, this means our algorithm's success in finding correspondence depends on the underlying generative model which generates its training data. Concern over GANs' tendency to drop modes is well-documented (e.g.,~\cite{bau2019seeing}). However, as latent variable generative models continue to improve---including efforts to improve mode coverage---we expect that GANgealing will improve as well.

\section{Assets}

\begin{itemize}
    \item \href{https://www.google.com/search?q=batmask&tbm=isch&tbs=ic:trans&hl=en-US&sa=X&ved=0CAMQpwVqFwoTCOimgqjFsfQCFQAAAAAdAAAAABAC&biw=1680&bih=829#imgrc=kXCmc5K08s5WeM}{Cat Batman Mask}
    \item \href{https://stock.adobe.com/search?load_type=search&native_visual_search=&similar_content_id=&is_recent_search=&search_type=usertyped&k=horse+soldier&asset_id=296994761}{Bird Soldier}
    \item \href{https://www.google.com/search?q=rudolph\%20nose&tbm=isch&tbs=ic:trans&hl=en-US&sa=X&ved=0CAMQpwVqFwoTCMC9_PLFsfQCFQAAAAAdAAAAABA0&biw=1680&bih=829#imgrc=Xx5uj9F40lYzvM}{Rudolph Dog}
    \item \href{https://www.deviantart.com/toxicsquall/art/Azumarill-water-Pokemon-tattoo-677312260}{CelebA Pokémon Tattoo 1}
    \item \href{https://www.pinterest.com/pin/309552174369121681/}{CelebA Pokémon Tattoo 2}
    \item \href{https://www.partycity.com/glitter-rainbow-unicorn-horn-793076.html}{Unicorn Horn}
    \item \href{https://schleese.com/product/devin-western-trail-saddle/}{Horse Saddle}
    \item \href{https://www.tenstickers.com/stickers/dragon-race-car-sticker-A18555}{Car Dragon Decal}

\end{itemize}